\documentclass[lettersize,journal]{IEEEtran}
\usepackage{amsmath,amsfonts}
\usepackage{algorithm}
\usepackage{algorithmic}
\usepackage{array}
\usepackage[caption=false,font=normalsize,labelfont=rm,textfont=rm]{subfig}
\usepackage{textcomp}
\usepackage{stfloats}
\usepackage{url}
\usepackage{verbatim}
\usepackage{graphicx}
\usepackage{subfig}
\usepackage{cite}
\usepackage{booktabs}
\usepackage{multirow}
\usepackage{orcidlink}  
\usepackage{titlesec}
\titleformat{\section}{\large\bfseries}{\arabic{section}}{1em}{}
\titleformat{\subsection}{\normalsize\bfseries}{\arabic{section}.\arabic{subsection}}{1em}{}
\titleformat{\subsubsection}{\normalsize\bfseries}{\arabic{section}.\arabic{subsection}.\arabic{subsubsection}}{1em}{}

\begin{document}

\title{kFuse: A novel density based agglomerative clustering}

\author{Huan Yan and Junjie Hu


\thanks{Huan Yan is with the School of Computer Science, Shaanxi Normal University, Xi'an 710119, China (e-mail: yan-huan@snnu.edu.cn).}
\thanks{Junjie Hu is with the Department of Computer Science and Engineering, Shanghai Jiao Tong University, Shanghai 200240, China (e-mail: nakamoto@sjtu.edu.cn).}}



\maketitle

\begin{abstract}
Agglomerative clustering has emerged as a vital tool in data analysis due to its intuitive and flexible characteristics. However, existing agglomerative clustering methods often involve additional parameters for sub-cluster partitioning and inter-cluster similarity assessment. This necessitates different parameter settings across various datasets, which is undoubtedly challenging in the absence of prior knowledge. Moreover, existing agglomerative clustering techniques are constrained by the calculation method of connection distance, leading to unstable clustering results. To address these issues, this paper introduces a novel density-based agglomerative clustering method, termed kFuse. kFuse comprises four key components: (1) sub-cluster partitioning based on natural neighbors; (2) determination of boundary connectivity between sub-clusters through the computation of adjacent samples and shortest distances; (3) assessment of density similarity between sub-clusters via the calculation of mean density and variance; and (4) establishment of merging rules between sub-clusters based on boundary connectivity and density similarity. kFuse requires the specification of the number of clusters only at the final merging stage. Additionally, by comprehensively considering adjacent samples, distances, and densities among different sub-clusters, kFuse significantly enhances accuracy during the merging phase, thereby greatly improving its identification capability. Experimental results on both synthetic and real-world datasets validate the effectiveness of kFuse.
\end{abstract}

\begin{IEEEkeywords}
Clustering, Agglomeration Clustering, Hierarchical Clustering, Natural Neighbors.
\end{IEEEkeywords}

\section{Introduction}
\IEEEPARstart{C}{lustering} is an indispensable tool in data analysis and pattern recognition\cite{ClusteringOnDataMining2002BF}, aimed at partitioning similar objects into distinct clusters to extract valuable insights from each group\cite{ClusteringUsage2014SS}. Its applications span various fields, including image processing\cite{ClusteringOnImage2020LT}, document classification\cite{ClusteringOnDocument2017MJP}, gene sequencing\cite{ClusteringOnGene2021LHF}\cite{ClusteringOnGene2023BHI}, and cybersecurity\cite{ClusteringOnCyber2016BAL}. Existing clustering algorithms can be broadly categorized into partition-based, hierarchical, graph-based, and density-based methods\cite{ClusteringAlgorithm1988JAK}.

Partition-based clustering algorithms assign samples to different sub-clusters. K-means\cite{Kmeans1966MJB} is a quintessential representative of partition-based clustering, which identifies K cluster centroids based on input data and assigns each point to the nearest centroid. While K-means is characterized by its simplicity and efficiency, it fails to recognize non-spherical clusters.

Hierarchical clustering algorithms distinguish different clusters by generating a hierarchical structure of clusters. Slink\cite{SLINK1973SR, SLINK1973RS1989HKS} and Chameleon\cite{Chameleon1999KG} are notable examples of hierarchical clustering algorithms that merge samples by calculating the connectivity among distinct samples to complete the clustering process. These algorithms can identify non-spherical clusters but are highly susceptible to noise and high-dimensional data.

Graph-based clustering algorithms treat data samples as points in a graph and partition these points into different clusters using unsupervised methods\cite{GraphedClustering2023ZH}. Spectral Clustering (SC) \cite{SC2000SJB,SC2007VL}is a prominent example, modeling the similarity relationships among data points as weights in a graph, and subsequently utilizing the eigenvalues and eigenvectors of the graph's Laplacian matrix for data segmentation and clustering. SC is particularly effective for clustering sparse data; however, it is heavily dependent on the similarity matrix, leading to potentially divergent clustering results with different matrices. Affinity Propagation (AP) \cite{AP2007BJF}is another classic graph-based clustering algorithm that constructs clusters by sending messages between data points until convergence. Unlike SC, AP does not require the number of clusters to be predetermined. Nonetheless, AP performs poorly on non-spherical clusters and has a higher computational complexity.

Density-based clustering algorithms compute the density of samples and iteratively connect adjacent high-density points to form distinct clusters. DBSCAN\cite{DBSCAN1996EM} is a classic density-based clustering algorithm that iteratively clusters different samples based on a specified radius, $\epsilon$. Although DBSCAN can identify arbitrary-shaped clusters, it requires two parameters and is highly sensitive to the $\epsilon$ parameter. OPTICS\cite{OPTICS1999AM}, as an extension of DBSCAN, eliminates the necessity for the $\epsilon$ parameter and produces a hierarchical result. However, the output of OPTICS is an order of distances rather than intuitive clusters, resulting in poorer interpretability. Density Peak Clustering (DPC)\cite{DPC2014AR}, a novel density-based clustering algorithm published in Science in 2014, allows for manual selection of suitable cluster centers and completion of clustering without prior knowledge. DPC can identify arbitrary-shaped clusters without iterative processes; however, its reliance on decision graphs for selecting cluster centers necessitates human involvement, introducing a degree of randomness. Furthermore, DPC exhibits poor fault tolerance, often resulting in a "domino effect," where an incorrectly clustered data object leads to misallocation of other connected objects.

Despite the notable achievements of the aforementioned methods, certain challenges remain. Inspired by agglomerative clustering, this paper proposes a novel density-based agglomerative clustering algorithm, kFuse. kFuse is a simple, efficient, and non-iterative clustering algorithm, with the following key contributions:

\begin{enumerate}
\item{A method for partitioning sub-clusters with reliable relationships without requiring any parameters for the initial partitioning of the original dataset.}
\item{A metric for assessing inter-cluster similarity to reliably evaluate the among sub-clusters.}
\item{The clustering process of kFuse is straightforward, comprising only two steps: partitioning and merging of sub-clusters}
\item{kFuse requires only a single parameter to estimate the number of clusters as input and possesses the capability to identify clusters of arbitrary shapes.}
\end{enumerate}

The remainder of the paper is organized as follows: Section 2 introduces related work; Section 3 presents the proposed method; Section 4 showcases experiments and discussions; and Section 5 concludes the study.

\section{Related works}

\subsection{Agglomerative clustering}

Agglomerative clustering constructs a clustering tree (dendrogram) in a bottom-up manner, starting from each sample point and progressively merging the two closest clusters until a single cluster encompassing all data points is formed\cite{ClusteringAlgorithm1999JAK}. Chameleon\cite{Chameleon1999KG} is a representative agglomerative clustering algorithm, which divides the clustering process into two steps:

\begin{enumerate}
\item{Initial Sub-cluster Partitioning: In this phase, the Chameleon algorithm constructs a graph using the $k$-nearest neighbors algorithm to connect each point in the dataset to its $k$ nearest neighbors, with each edge corresponding to a weight. Subsequently, Chameleon employs a graph partitioning technique to segment the initial graph, yielding distinct regions with minimal edge weights.}
\item{Dynamic Merging of Sub-clusters: In this phase, the Chameleon algorithm assesses the similarity between different clusters by calculating their relative inter connectivity (RI) and relative closeness (RC), using these metrics as criteria for merging sub-clusters until no further merge conditions are met, or only one cluster remains.}
\end{enumerate}

The Chameleon algorithm is characterized by high adaptability and interpretability, capable of discovering clusters of arbitrary shapes, which has led to significant achievements. However, it involves numerous parameters, including the $k$ value in $k$-nearest neighbors, the metric $\alpha$, and the selection of thresholds. Furthermore, Chameleon exhibits poor performance on high-dimensional datasets and is vulnerable to outliers and noise.

From the execution flow of Chameleon, its primary advantage lies in its strategy of merging sub-clusters based on specific metrics, effectively circumventing the limitations of partition-based clustering techniques (such as K-means), which struggle to identify clusters of varying sizes and non-convex shapes. In this regard, the proposed kFuse also divides the clustering process into two main stages: sub-cluster partitioning and merging, enhancing its ability to recognize clusters of arbitrary shapes.

Notably, Chameleon’s sub-cluster partitioning relies on $k$-nearest neighbors, necessitating the provision of a $k$ parameter. To mitigate the reliance on hyperparameters, kFuse introduces the concept of natural neighbors.

\subsection{Natural neighbor}

Natural neighbors\cite{NaturalNeighbors2016QZ} is a novel concept of neighbor. It posits that if a data object $x$ considers $y$ as its neighbor, and $y$ similarly regards $x$ as its neighbor, then $y$ is one of the natural neighbors of $x$. Natural neighbors have shown substantial application value in outlier detection\cite{DataMiningBasedNN2016JH}, clustering analysis\cite{ClusteringBasedNN2022ZL}, and dataset augmentation\cite{SMOTEBasedNN2021SC}.

Let $D$ represent the dataset, $d(x,y)$ be the distance between points $x$ and $y$, and point $o$ be the $k$-th nearest neighbor of point $p$. The definitions of $k$-nearest neighbors and reverse $k$-nearest neighbors can be articulated as follows:

\noindent {\bf{Definition 1 ($k$-nearest neighbors) }}. The $k$-nearest neighbors of point $p$ form a set of points that satisfy the following condition: $NN_k(p)=\{x\in D \mid d(p,x) \leq d(p,o)\}$.

\noindent {\bf{Definition 2 (Reverse $k$-nearest neighbors)}}. The reverse $k$-nearest neighbors of point $p$ form a set of points that satisfy the following condition: $RNN_k(p)=\{x\in D \mid p \in NN_k(x)\}$.

\begin{figure}[htbp]
\begin{algorithm}[H]
\caption{NaN-Searching}
\label{Alg_NaNSearch}
\begin{algorithmic}[1] 
	\REQUIRE{$D$ (the data set).}
	\ENSURE {$\lambda, nb$.} 
 
	\STATE Initializing: $r=1; Num(0)=size(D)$;
	\FOR{each data point $p \in D$} 
	{
        \STATE $nb(p)=0; NN_0(p)=\emptyset; RNN_0(p)=\emptyset;$
    }
    \ENDFOR
	\WHILE{true}
	{
		\FOR{each data point $p \in D$}
		{
			\STATE Find the $r$-th neighbor $q$ of $p$;
			\STATE $NN_r(p)=NN_{r-1}(p) \cup \{q\}$;
			\STATE $RNN_r(q)=RNN_{r-1}(q) \cup \{p\}$;
			\STATE $nb(q)=nb(q)+1$;
		}
        \ENDFOR
		\STATE $Num(r)=length(find(nb==0))$;
		\IF{$Num(r)==Num(r-1)$}
		{
			\STATE break;
		}
        \ENDIF
		\STATE $r=r+1$;
	}
    \ENDWHILE
	\STATE $\lambda=r$;
    \RETURN $\lambda, nb$;
 \end{algorithmic}
\end{algorithm}	
\end{figure}

To obtain the natural neighbors each point in the dataset, it is necessary to dynamically expand the search for neighbors among different samples. This process ultimately yields a critical search range known as the natural characteristic value, formally defined as follows:

\noindent {\bf{Definition 3 (Natural characteristic $\lambda$)}}. Starting from 1 the search range $r$ for neighbors is expanded. Once, in an iteration, the number of samples with no reverse neighbors remains unchanged, this search $r$ is identified as the natural characteristic value for the current dataset.

\noindent {\bf{Definition 4 (Natural neighbors) }}. The natural neighbors of sample $p$ form a set of points that satisfy the following condition:$\{x\in D \mid x\in NN_{\lambda}(p) \cap p\in NN_{\lambda}(x)\}$.

It is noteworthy that the computation of natural neighbors is parameter-free, meaning that upon inputting the dataset, the natural neighbors of each point are already established.

The principal contribution of natural neighbors to kFuse lies in providing a parameter-free approach to obtaining reliable neighbors among different samples, facilitating the computation of densities for various sample points and the initial partition of sub-clusters. Algorithm~\ref{Alg_NaNSearch} details the of acquiring the natural characteristic value, where $NN_r(p)$ represents the $r$-nearest neighbors point $p$, $RNN_r(q)$ denotes the reverse $r$-nearest neighbors of point $q$, and $nb(q)$ indicates the number of reverse neighbors for point $q$.

\section{The proposed kFuse algorithm}

\begin{figure*}[!t]
    \centering
    \includegraphics[width=0.95\textwidth]{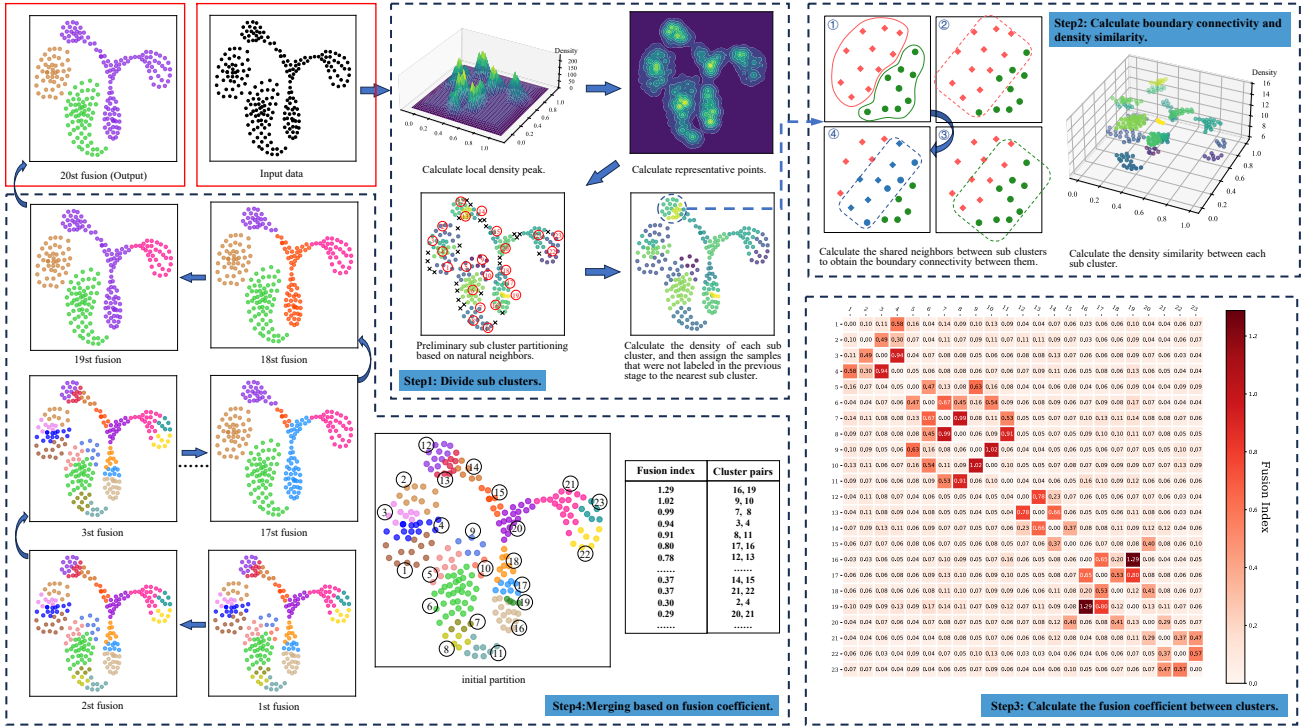}
    \caption{The clustering process of kFuse on the Away\cite{CustomDatasets} dataset.}
    \label{Fig_ClusteringProcess}
\end{figure*}

The kFuse algorithm adopts the basic framework of agglomerative clustering\cite{AHClustering2006GG}, where the dataset is first partitioned and then the resulting sub-clusters are merged. The Fig.~\ref{Fig_ClusteringProcess} illustrates the clustering process of kFuse on the Away\cite{CustomDatasets} dataset.

\subsection{Divide sub clusters}

Traditional agglomerative clustering begins by treating each sample as a separate cluster, merging the most similar clusters at each step. This significantly increases the number of subsequent merges. A simple yet effective solution is to group closely related points in the dataset into sub-clusters from the outset. This approach offers three key advantages:

\begin{enumerate}
\item{Reduces the number of cluster merges.}
\item{Improves the identification of clusters with complex structures.}
\item{Allows more information to be considered when evaluating inter-cluster similarity.}
\end{enumerate}

Building on this, kFuse leverages natural neighbors to achieve an initial partitioning of the dataset. In this step, kFuse first computes the local density of each sample using reverse nearest neighbors obtained by Algorithm~\ref{Alg_NaNSearch} and identifies local density peaks. These peaks are then expanded via natural neighbors to form initial partitions of the dataset.

\subsubsection{Local density peak}

To identify local density peaks, the density of each sample must first be calculated. In dense regions, the sum of distances between a sample and its nearest neighbors is typically smaller than in sparse regions, and samples in dense areas also tend to have more reverse nearest neighbors. Thus, a sample’s density is inversely proportional to its distance from neighboring points and directly proportional to the number of reverse nearest neighbors. The density kernel function of kFuse is defined as follows:

\begin{equation}
\rho(p)=\frac{nb(p)}{\sum_{q\in NN_k(p)}d(p,q)}
\label{Eq_density} 
\end{equation}
Where, $k$ is obtained by Algorithm~\ref{Alg_NaNSearch}, $NN_k(p)$ represents the $k$-nearest neighbor of point $p$, and $d(p,q)$ represents the distance between point $p$ and point $q$. From this, it is evident that kFuse computes densities without requiring any parameters, which is one of its key features.

After determining the densities, the local density peaks in the dataset can be identified as follows:

\noindent {\bf{Definition 5 (Representative) }}. Let $q$ be the point with the highest density among $p$ and its $k$ nearest neighbors. Then $q$ is the representative of $p$, denoted as $Rep(p) = q$.

\noindent {\bf{Definition 6 (Local density peak) }}. If $Rep(p) = p$, then $p$ is a local density peak.

\begin{figure}[htbp]
\centering
\subfloat[Away]{
\label{Fig_Representative_Away}
\includegraphics[width=0.46\columnwidth]{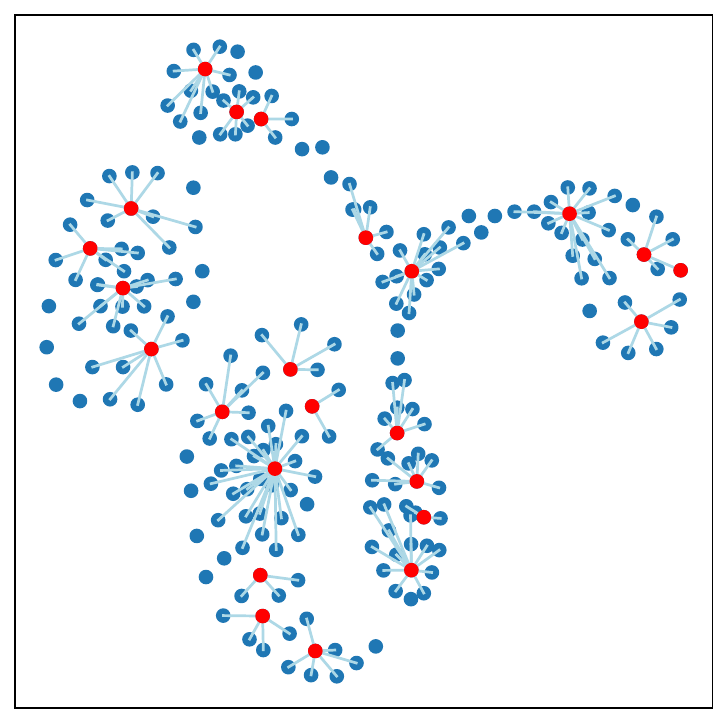}
}
\subfloat[Flame]{
\label{Fig_Representative_Flame}
\includegraphics[width=0.46\columnwidth]{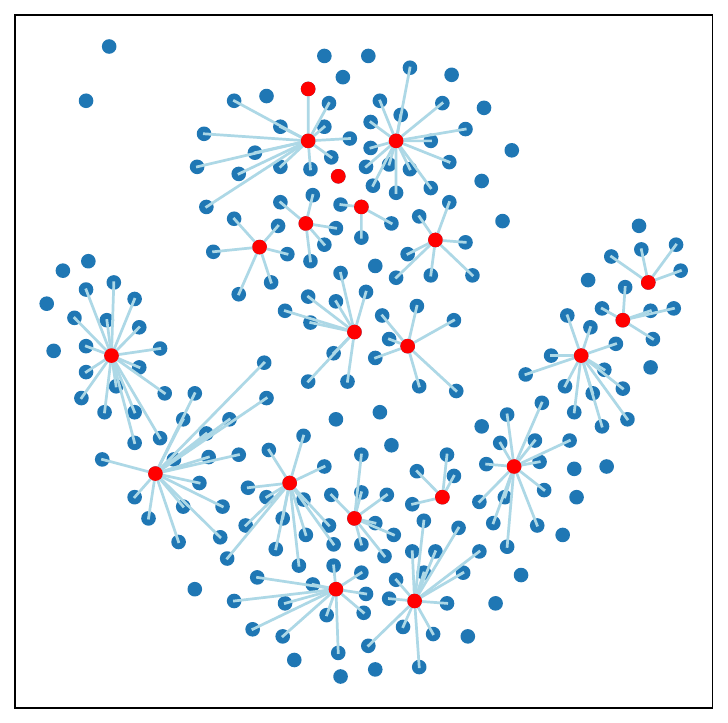}
}
\vskip 1pt
\subfloat[Zelnik1]{
\label{Fig_Representative_Zelnik1}
\includegraphics[width=0.46\columnwidth]{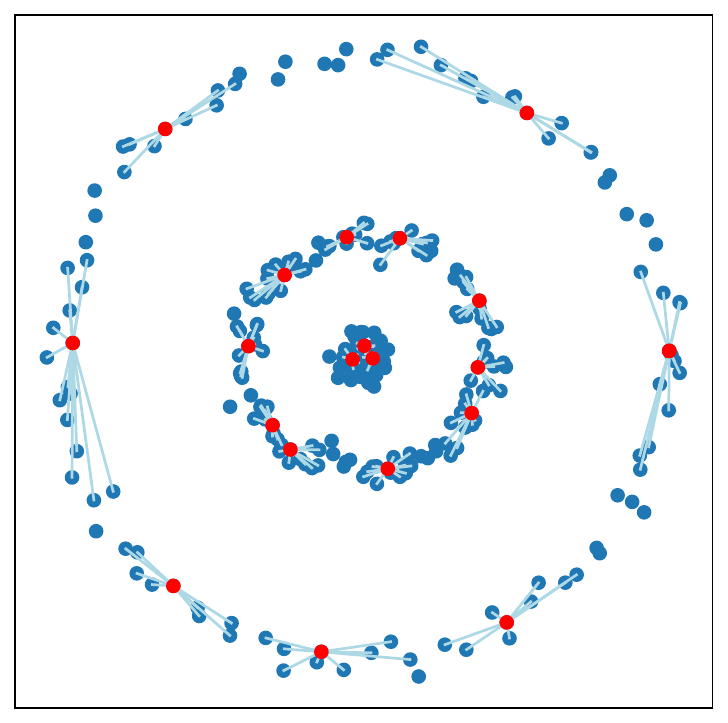}
}
\subfloat[Path-based]{
\label{Fig_Representative_Away}
\includegraphics[width=0.46\columnwidth]{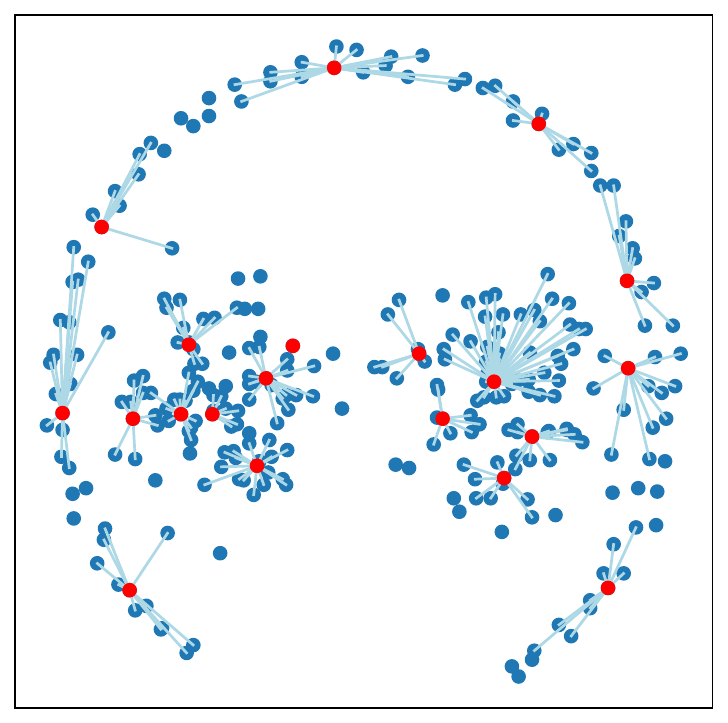}
}
\caption{Representative points on Away\cite{CustomDatasets}, Flame\cite{ClusteringDatasets}, Zelnik1\cite{ZelnikDatasets}, Path-based\cite{ClusteringDatasets} datasets.}
\label{Fig_Representative}
\end{figure}

Fig.~\ref{Fig_Representative} shows the identified local density peaks (red dots) in part of the dataset.

\subsubsection{Sub cluster partitioning strategy based on natural neighbors}

From Fig.~\ref{Fig_Representative}, it can be observed that the local density peaks and their surrounding natural neighbors have overall partitioned the original dataset. However, there are still shortcomings: in denser regions, several closely located local density peaks may emerge, which, due to the limitation of the $k$ value, are not merged into a single sub-cluster. To address this issue, we will merge these local density peaks (and their surrounding samples) based on their natural neighbors, thereby expanding the range of the partitioned sub-clusters.

Algorithm~\ref{Alg_DividingClusters} details the kFuse process for partitioning sub-clusters. Here, $Labels$ represent the label vector, where $Labels(p) = c$ indicates that sample $p$ belongs to cluster $c$. $S$ represents the set of sub-clusters, and $S_c$ denotes the sample set of cluster $c$.

\begin{figure}[!t]
\begin{algorithm}[H]
\caption{Dividing sub clusters}
\label{Alg_DividingClusters}
\begin{algorithmic}[1]
	\REQUIRE{$D$ (the data set), $\rho$, $NN_k$.}
	\ENSURE {$Labels,\ S$.} 
 
	\STATE Initializing: $labelCount=1, Labels=\{0\}, Queue \ Q, S = \emptyset$;
	\FOR{each data point $p \in D$} 
	{
        \STATE $Rep(p)=\mathop{\arg\max}\limits_{q\in \{NN_k(p)\cup p\}}(\rho(q));$
    }
    \ENDFOR
    \STATE $Rs=\{Rep\}$
	\FOR{each representative point $r \in Rs$}
	{
        \IF{$Labels(r)==0$}
        {
            \STATE $Labels(r) = labelCount$;
			\STATE $LabelCount = labelCount+1$;
			\STATE $Q.push(r)$;
        }
        \ENDIF
		\WHILE{$Q$ is not empty}
		{
			\STATE $head = Q.pop()$;
            \FOR{each neighbor $n \in NN_k(head)$}
            {
                \IF{$Labels(n)==0$ and $head \in NN_k(n)$}
                {
                    \STATE $Labels(n) = Labels(head)$;
                    \IF{$n \in Rs$}
                    {
                        \STATE $Q.push(n)$;
                    }
                    \ENDIF
                }
                \ENDIF
            }
            \ENDFOR
		}
        \ENDWHILE
	}
    \ENDFOR
	\STATE $c = 1$;
    \WHILE{$c < labelCount$}
    {
        \STATE $S_c=\{p\in D \mid Labels(p)=c\}$;
		\STATE $c = c+1$;
    }
    \ENDWHILE
    \RETURN $Labels, S$;
 \end{algorithmic}
\end{algorithm}	
\end{figure}

Note that after Algorithm~\ref{Alg_DividingClusters} finishes, some outliers may remain unlabeled (i.e., not assigned to any sub-cluster). This is intentional, allowing for the exclusion of outliers when calculating the density of each sub-cluster. Once the sub-cluster densities are computed, kFuse assigns the unlabeled samples to the nearest sub-cluster, thus completing the initial partitioning.

\subsection{Fusion sub cluster}

This section proposes a cluster fusing index based on boundary connectivity and density similarity, which reliably assesses the cohesion between sub-clusters, enabling the fusion of sub-clusters from the previous stage to complete the clustering process.

\subsubsection{Boundary connectivity}

Considering that Euclidean distance is not an ideal metric for manifold data and that precise geodesic distance calculations are computationally expensive, \cite{LDPMST2021CDD} proposed a local density peak distance based on shared neighbors. Inspired by this, we define a new inter-cluster similarity metric called boundary connectivity by calculating adjacent samples and shortest distances between different sub-clusters.

To find the adjacent samples between two sub-clusters, the sub-clusters must first be augmented.

\noindent {\bf{Definition 7 (Augmented sub-cluster) }}. For any sub-cluster $S_c$, its augmented sub-cluster $AS_c$ is the union of all its members and their $\lambda$ nearest neighbors, where $\lambda$ is a natural characteristic value. Formally:

\begin{equation}
AS_c=S_c \cup \{\cup_{p\in S_c}NN_{\lambda}(p)\}
\label{Eq_AS} 
\end{equation}

\noindent {\bf{Definition 8 (Adjacent samples) }}. The adjacent samples of sub-clusters $S_i$ and S$_j$, denoted as $ADS_{i,j}$, are the intersection of their respective augmented sub-clusters. Formally:

\begin{equation}
ADS_{i,j}= AS_i \cap AS_j
\label{Eq_ADS} 
\end{equation}

\begin{figure}[htbp]
\centering
\subfloat[]{
\label{Fig_ADS1}
\includegraphics[width=0.46\columnwidth]{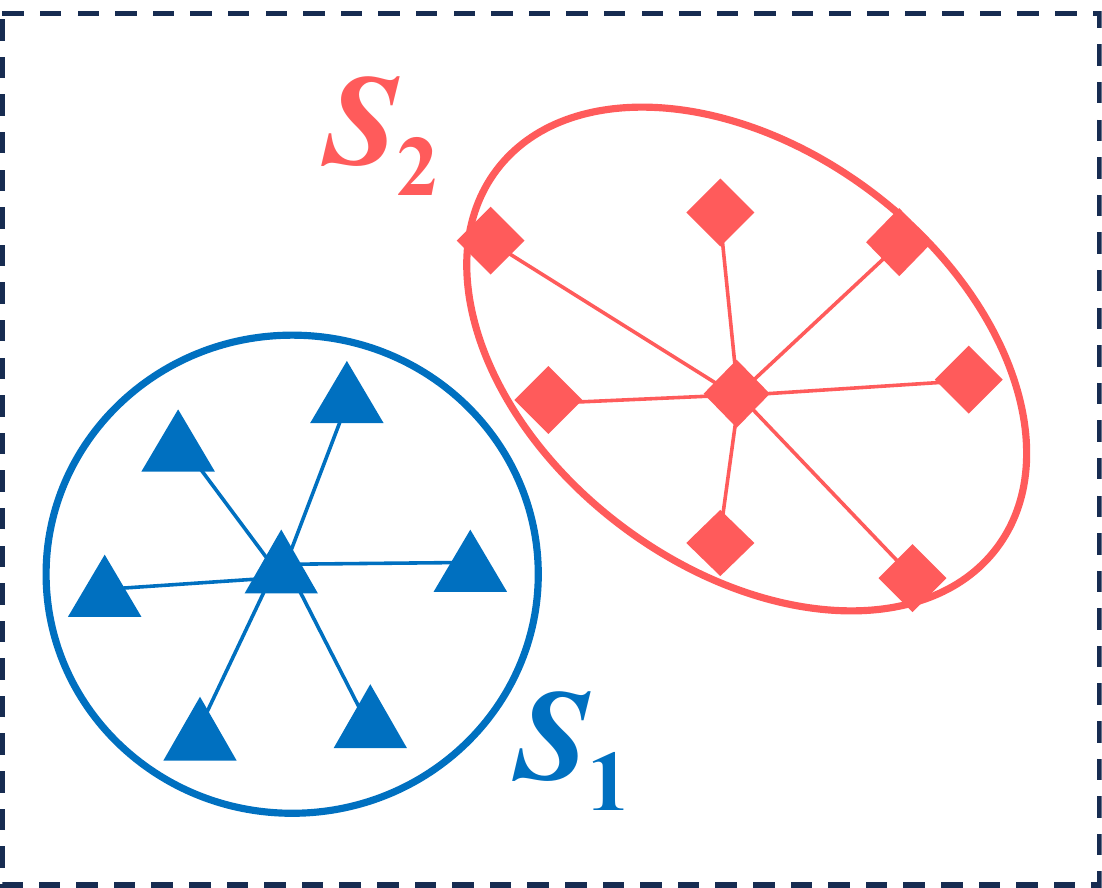}
}
\subfloat[]{
\label{Fig_ADS2}
\includegraphics[width=0.46\columnwidth]{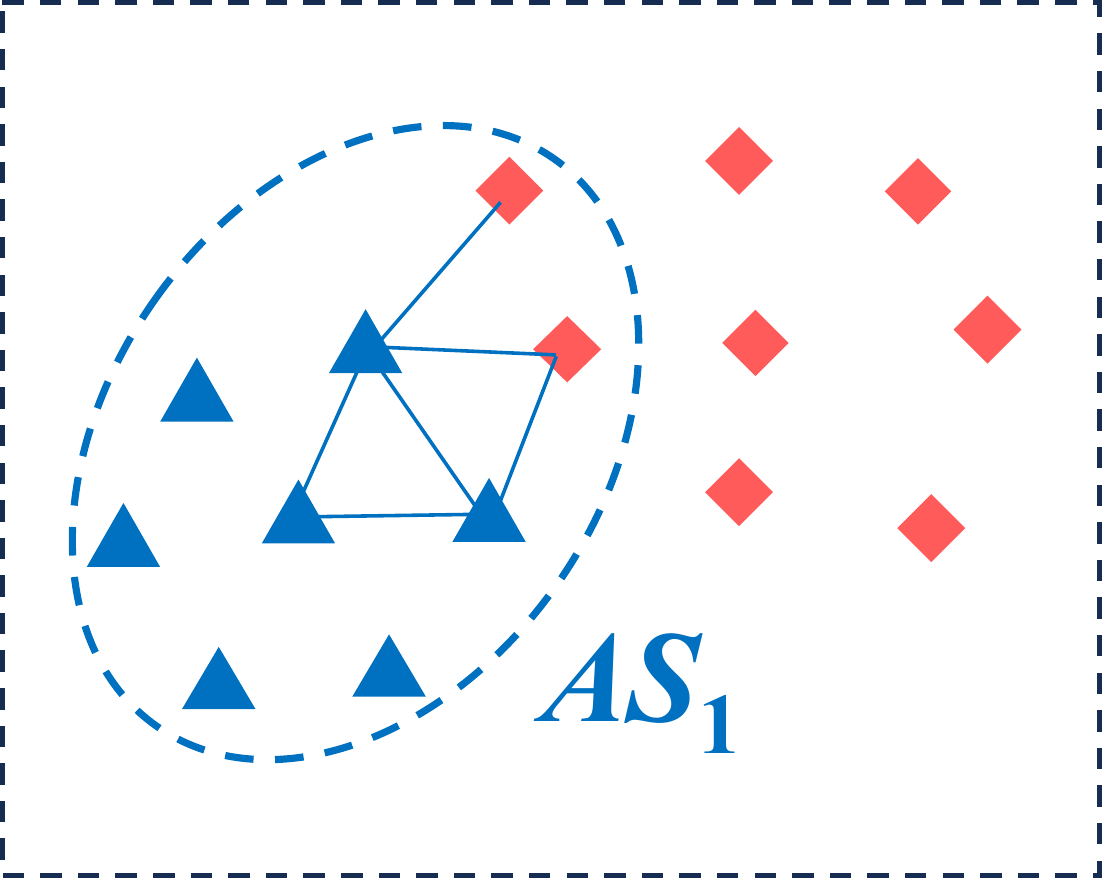}
}
\vskip 1pt
\subfloat[]{
\label{Fig_ADS3}
\includegraphics[width=0.46\columnwidth]{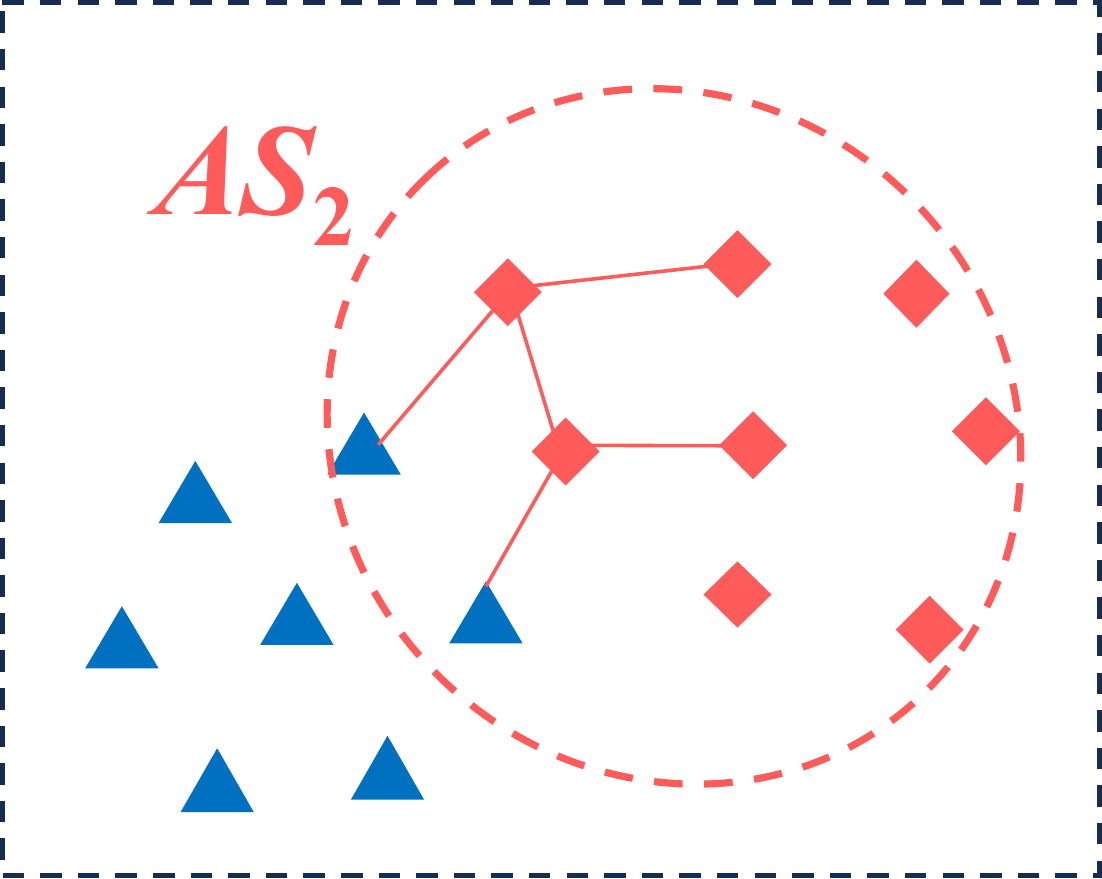}
}
\subfloat[]{
\label{Fig_ADS4}
\includegraphics[width=0.46\columnwidth]{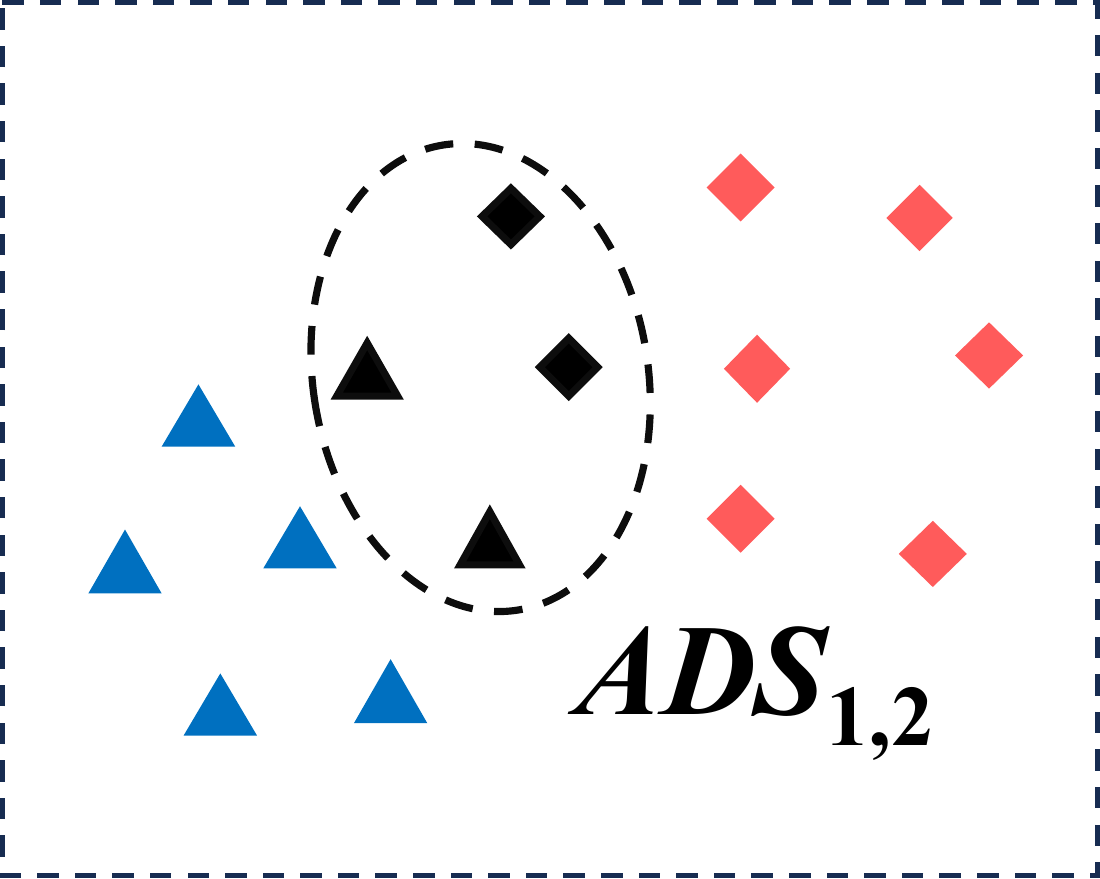}
}
\caption{Adjacent samples between two clusters. (a) Groudh truth of $S_1$ and $S_2$. (b) Augmented sub-cluster of $S_1$. (c) Augmented sub-cluster of $S_2$. (d) Adjacent samples of $S_1$ and $S_2$.}
\label{Fig_ADS}
\end{figure}

The adjacency samples between sub clusters reflect the degree of adjacency between these two sub clusters, providing a reliable basis for measuring the boundary connectivity of these two sub clusters. Fig.~\ref {Fig_ADS} shows the detailed process of calculating the adjacency samples of sub clusters $S_1$ and $S_2$.

\begin{figure}[htbp]
\centering
\subfloat[]{
\label{Fig_DC1}
\includegraphics[width=0.29\columnwidth]{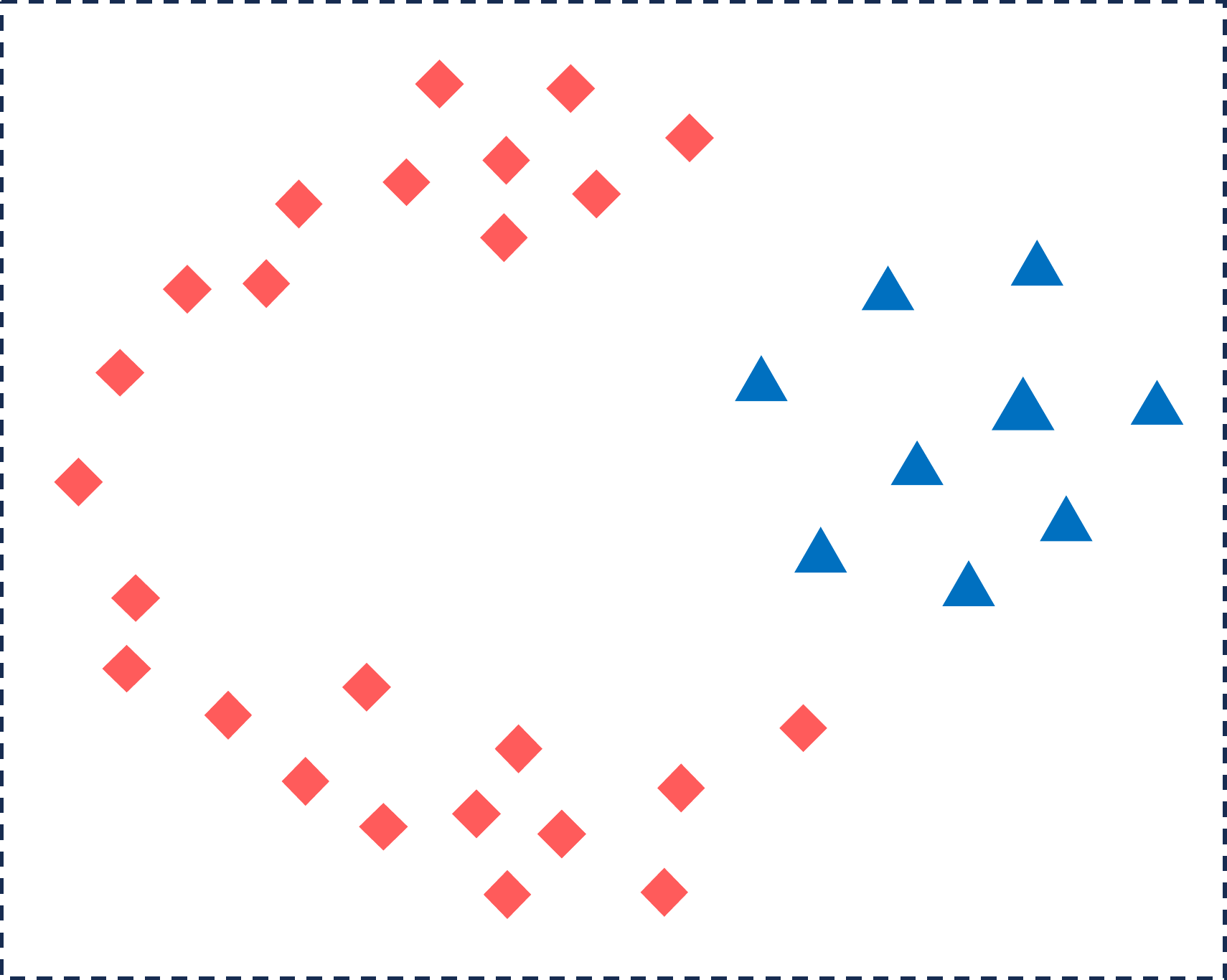}
}
\subfloat[]{
\label{Fig_DC2}
\includegraphics[width=0.29\columnwidth]{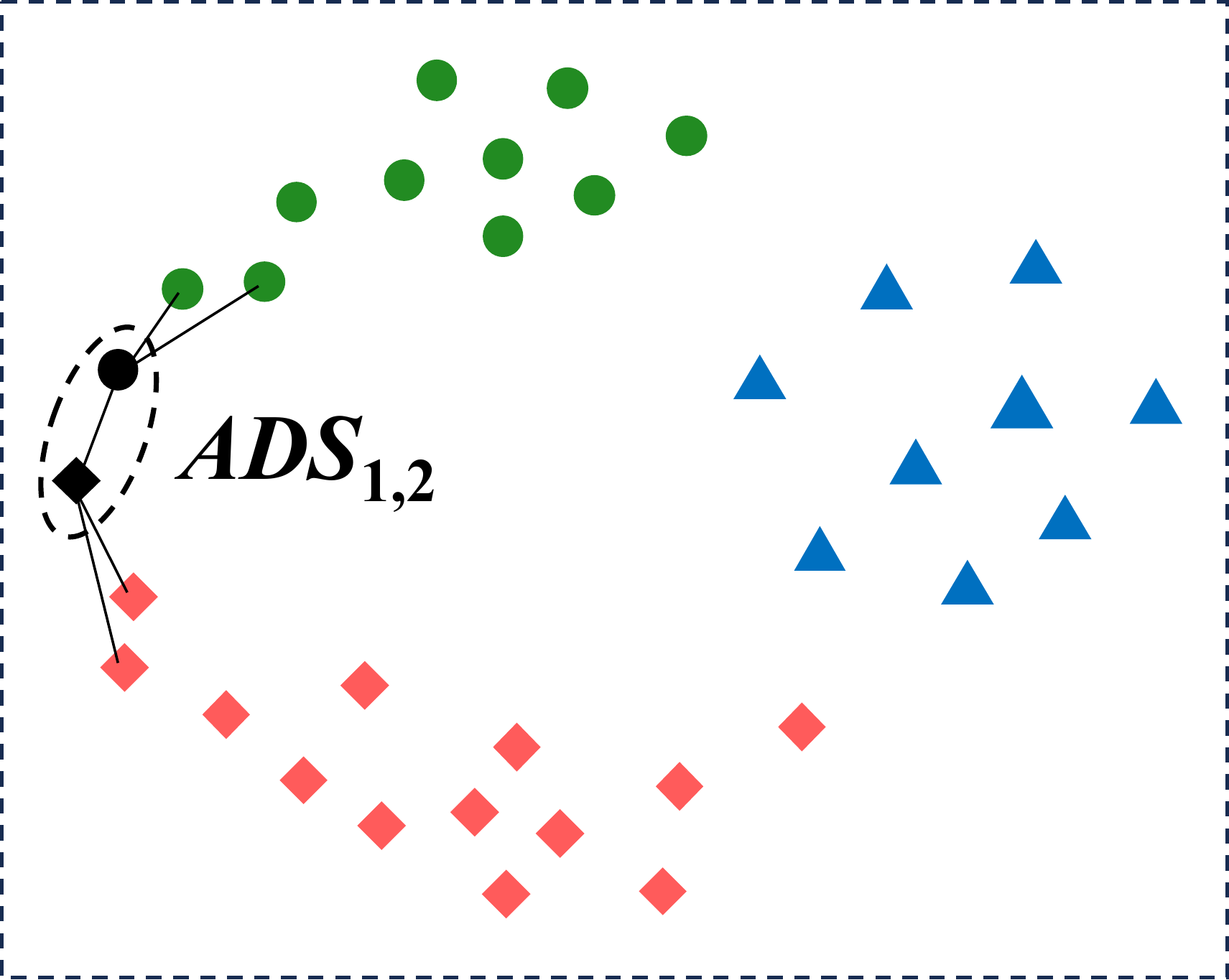}
}
\subfloat[]{
\label{Fig_DC3}
\includegraphics[width=0.29\columnwidth]{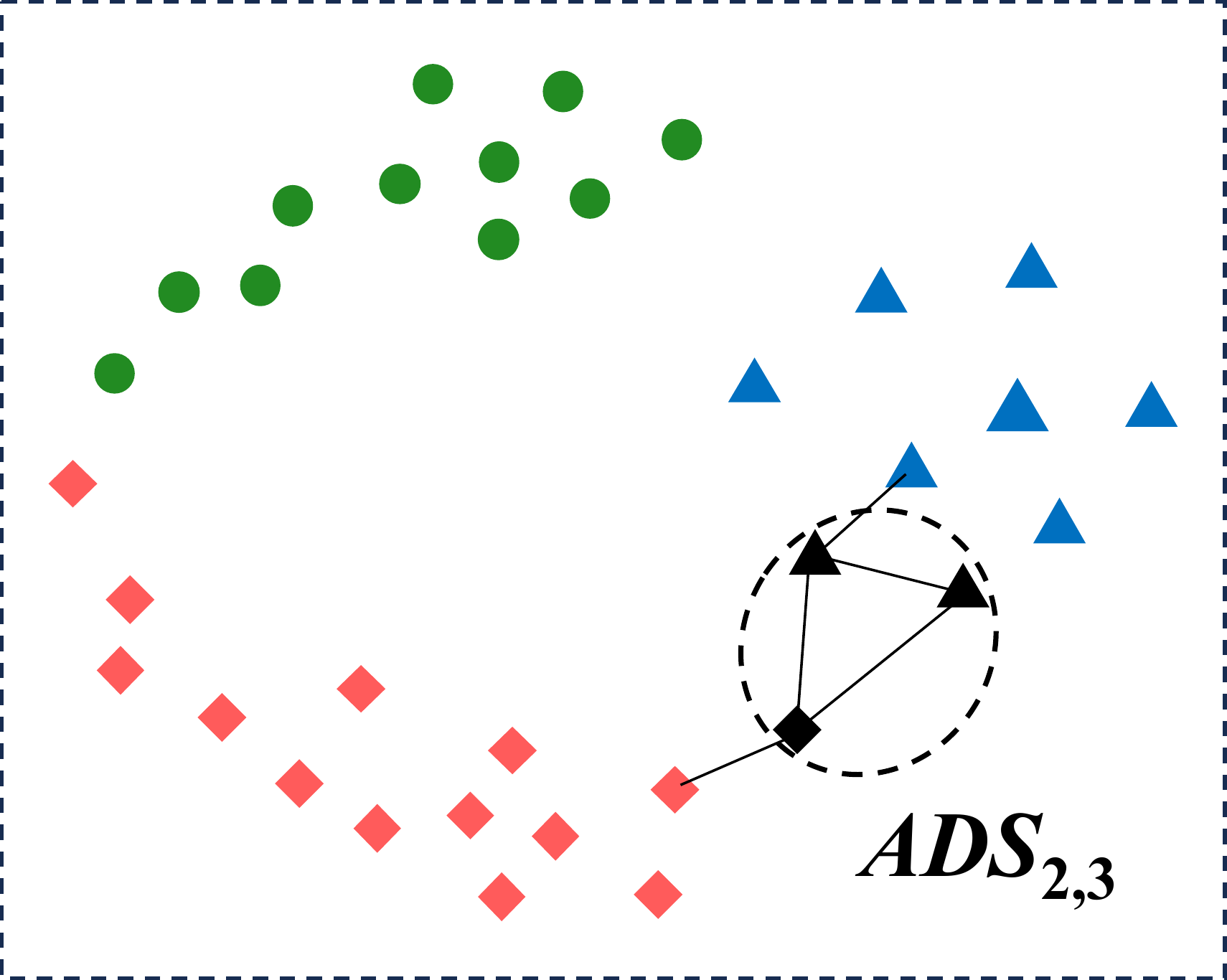}
}
\vskip 1pt
\subfloat[]{
\label{Fig_DC4}
\includegraphics[width=0.29\columnwidth]{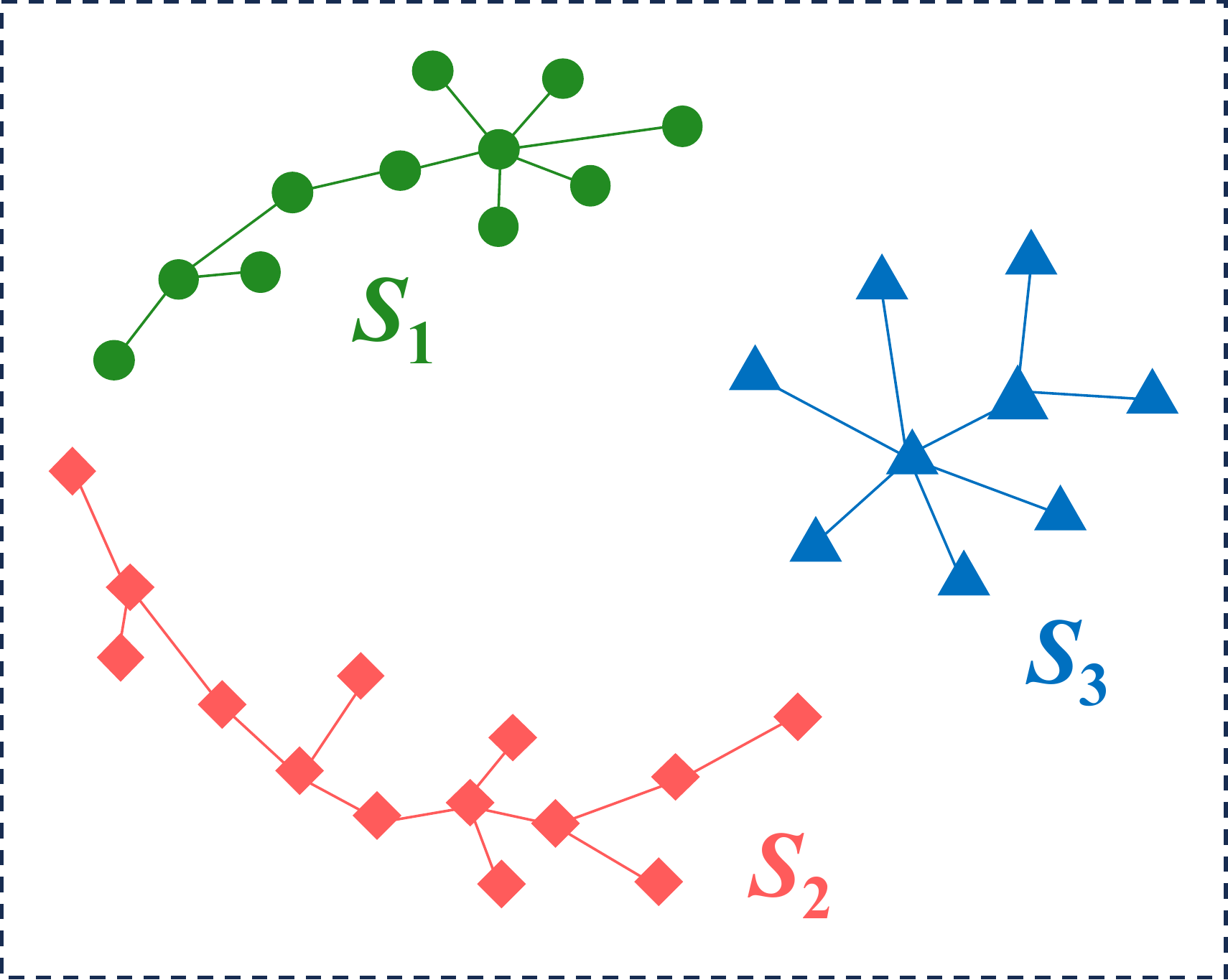}
}
\subfloat[]{
\label{Fig_DC5}
\includegraphics[width=0.29\columnwidth]{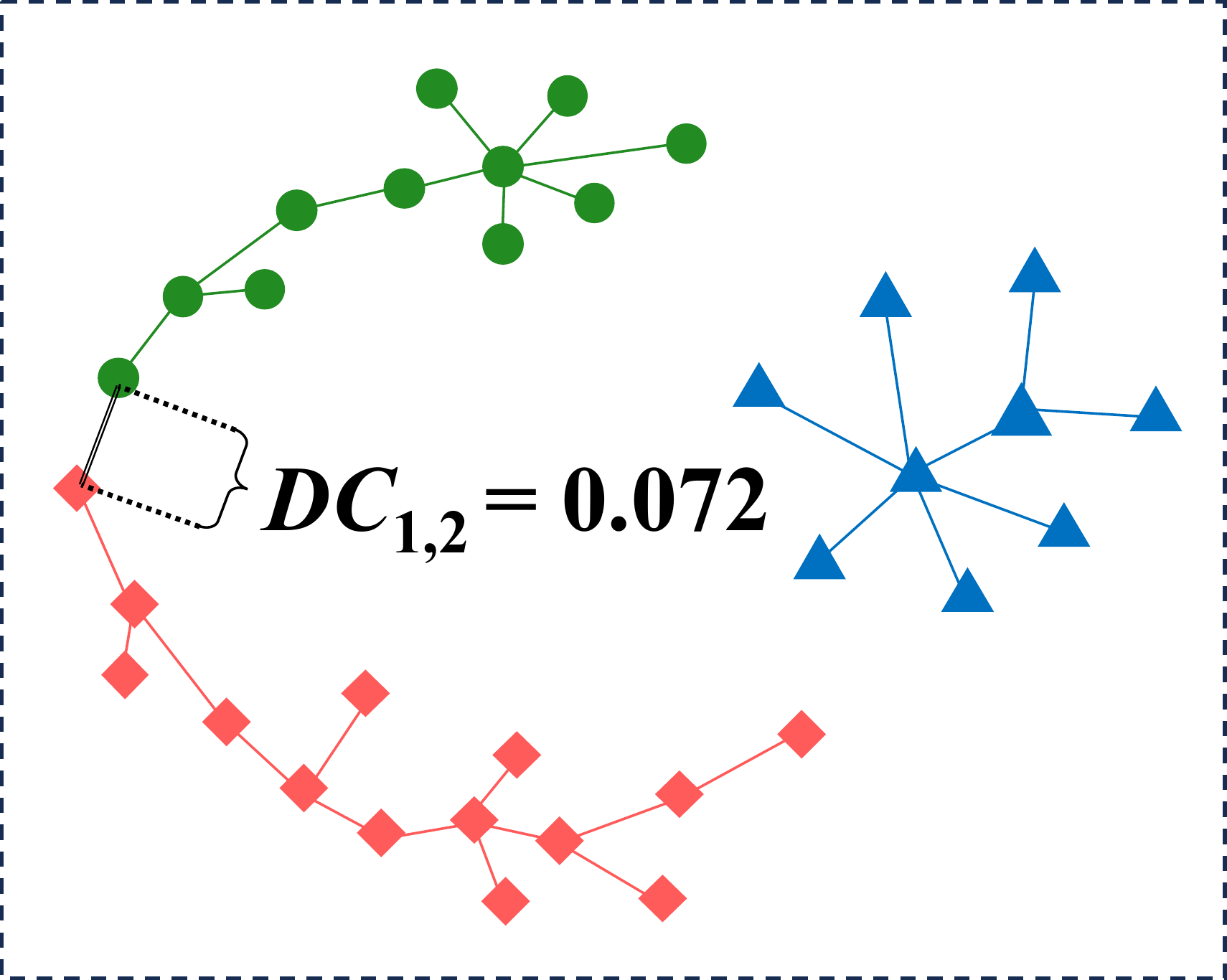}
}
\subfloat[]{
\label{Fig_DC6}
\includegraphics[width=0.29\columnwidth]{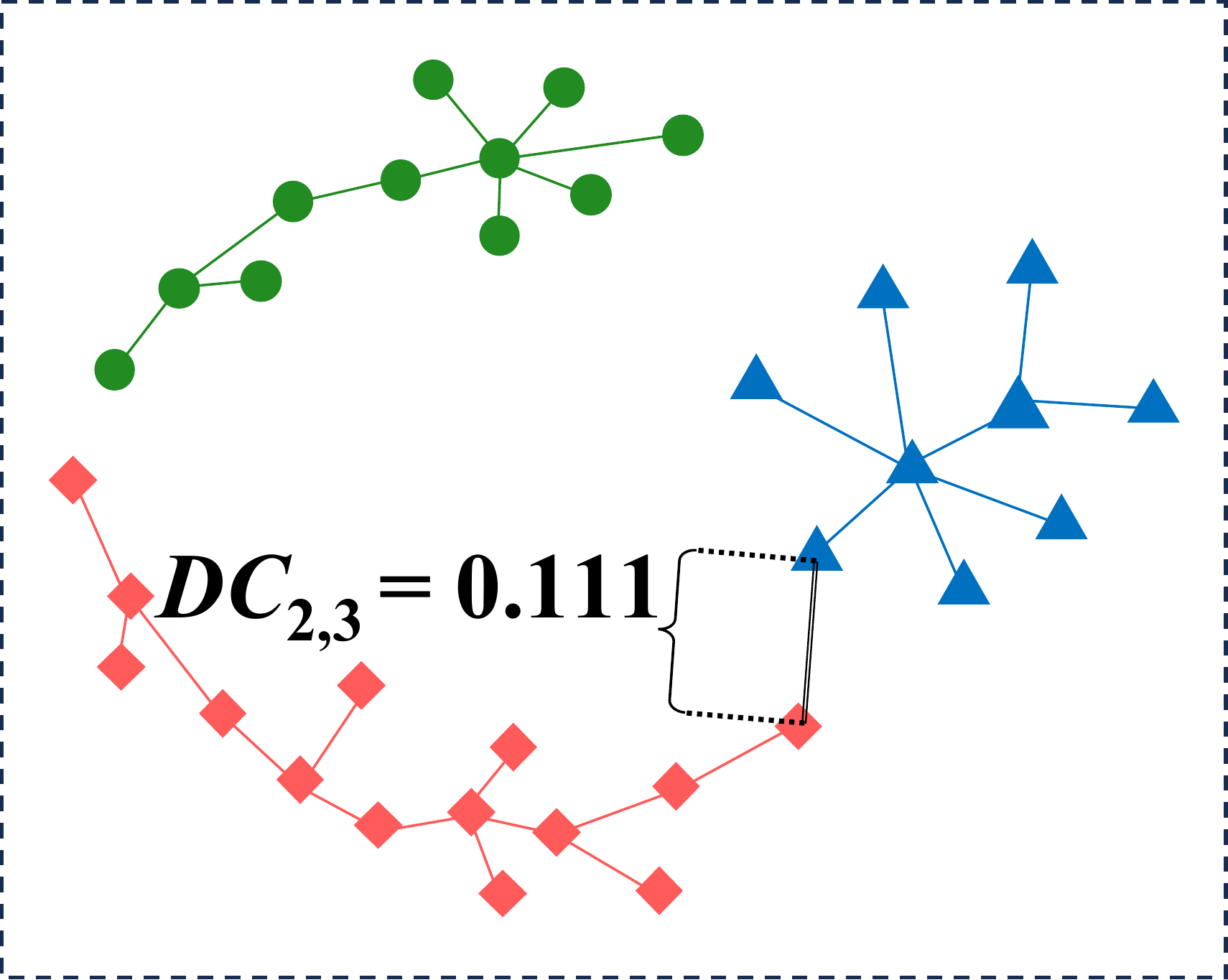}
}
\caption{The contribution of inter cluster distance to boundary connectivity. (a) The ground truth of dataset. (b) Adjacent samples between $S_1$ and $S_2$. (c) Adjacent samples between $S_2$ and $S_3$. (d) Initial partitioning of kFuse on the dataset. (e) minimum distance between $S_1$ and $S_2$. (f) minimum distance between $S_2$ and $S_3$.}
\label{Fig_DC}
\end{figure}

For complex-structured sub-clusters, the difference in adjacent samples between clusters may be minimal. Additionally, closer sub-clusters may sometimes have fewer adjacent samples (Fig.~\ref{Fig_DC} illustrates this case). Therefore, evaluating boundary connectivity based solely on adjacent samples is insufficient. To address this, kFuse introduces inter-cluster distance. Using the Single Link method\cite{SINGLELINK1973SR}, the inter-cluster distance $DC_{i,j}$ between sub-clusters $S_i$ and $S_j$ is defined as the minimum distance between any two samples from $S_i$ and $S_j$. Formally:

\begin{equation}
DC_{i,j}= \min_{p\in S_i, q\in S_j}d(p,q)
\label{Eq_DC} 
\end{equation}

Inter-cluster distance reflects the proximity of two sub-clusters, complementing the limitations of adjacent samples in manifold datasets.

Figure Fig~\ref{Fig_DC} shows that $S_1$ and $S_2$ have two adjacent samples, while $S_2$ and $S_3$ have three. If boundary connectivity were evaluated solely by adjacent samples, the connectivity between $S_2$ and $S_3$ would be higher than between $S_1$ and $S_2$. However, this is counterintuitive, as Fig.~\ref{Fig_DC5} and Fig.~\ref{Fig_DC6} show that the minimum distance between $S_1$ and $S_2$ is 0.072, while that between $S_2$ and $S_3$ is 0.111. Clearly, $S_1$ and $S_2$ are closer, and their boundary connectivity should be higher. This demonstrates that inter-cluster distance is a crucial reference when evaluating boundary connectivity.

Equation~\ref{Eq_Connectivity} provides the detailed formula for calculating the boundary connectivity $Con_{i,j}$ between sub-clusters $S_i$ and $S_j$:

\begin{equation}
Con_{i,j}= \frac{(ADS_{i,j}+1)\times e^{(-DC_{i,j})}}{\min\left(\lVert S_i\rVert, \lVert S_j\rVert\right)}
\label{Eq_Connectivity} 
\end{equation}
Where, $\lVert S_c\rVert$ denotes the number of samples in cluster $S_c$, and the constant 1 prevents boundary connectivity from reaching 0 between sub-clusters with no adjacent samples.

From Equation~\ref{Eq_Connectivity}, it is evident that boundary connectivity $Con_{i,j}$ increases with more adjacent samples and closer distances between sub-clusters. $Con_{i,j}$ reflects the spatial proximity of $S_i$ and $S_j$, which accurately captures the cohesion between sub-clusters in most cases. 

\begin{figure}[htbp]
\centering
\subfloat[]{
\label{Fig_DefectsInConnectivity1}
\includegraphics[width=0.29\columnwidth]{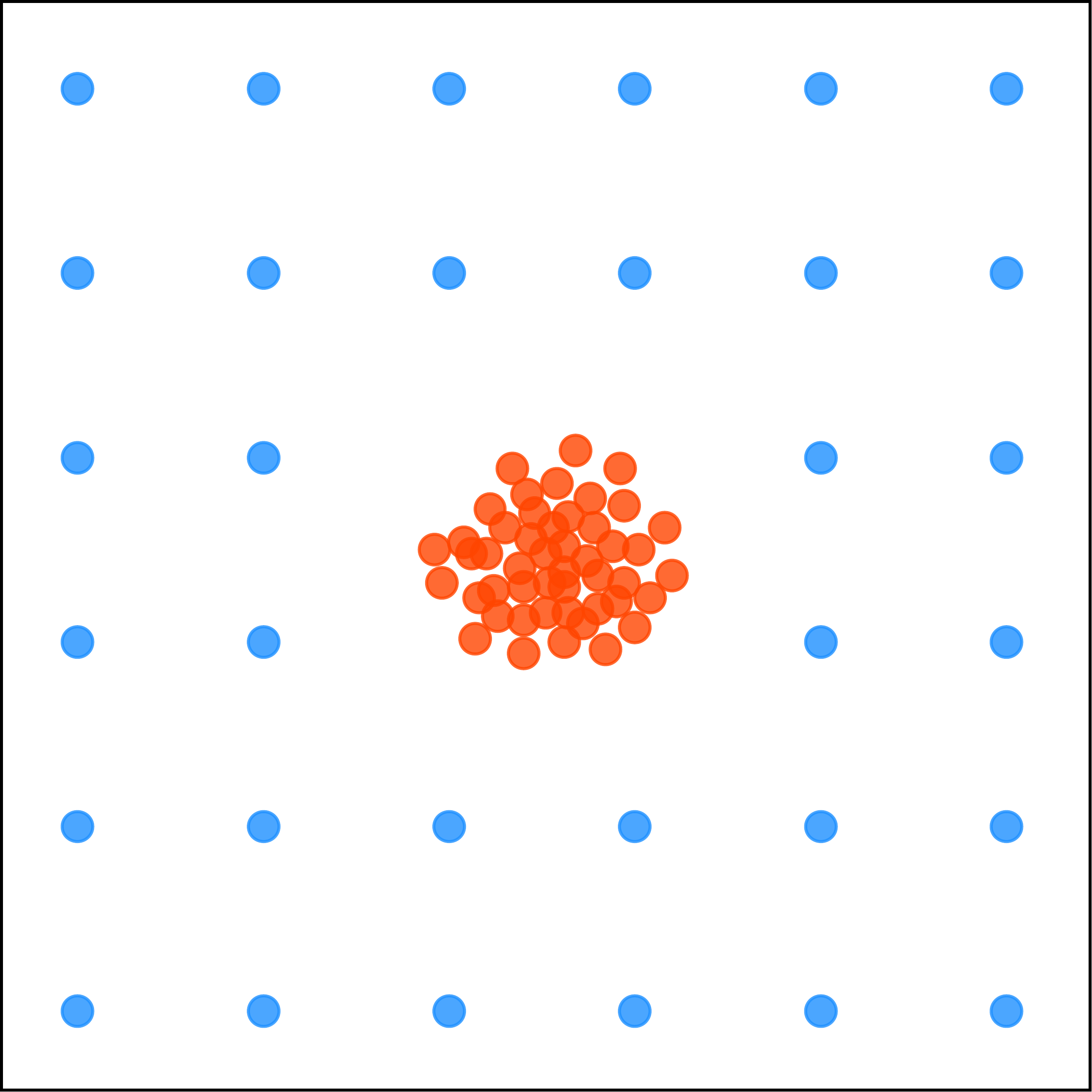}
}
\subfloat[]{
\label{Fig_DefectsInConnectivity2}
\includegraphics[width=0.29\columnwidth]{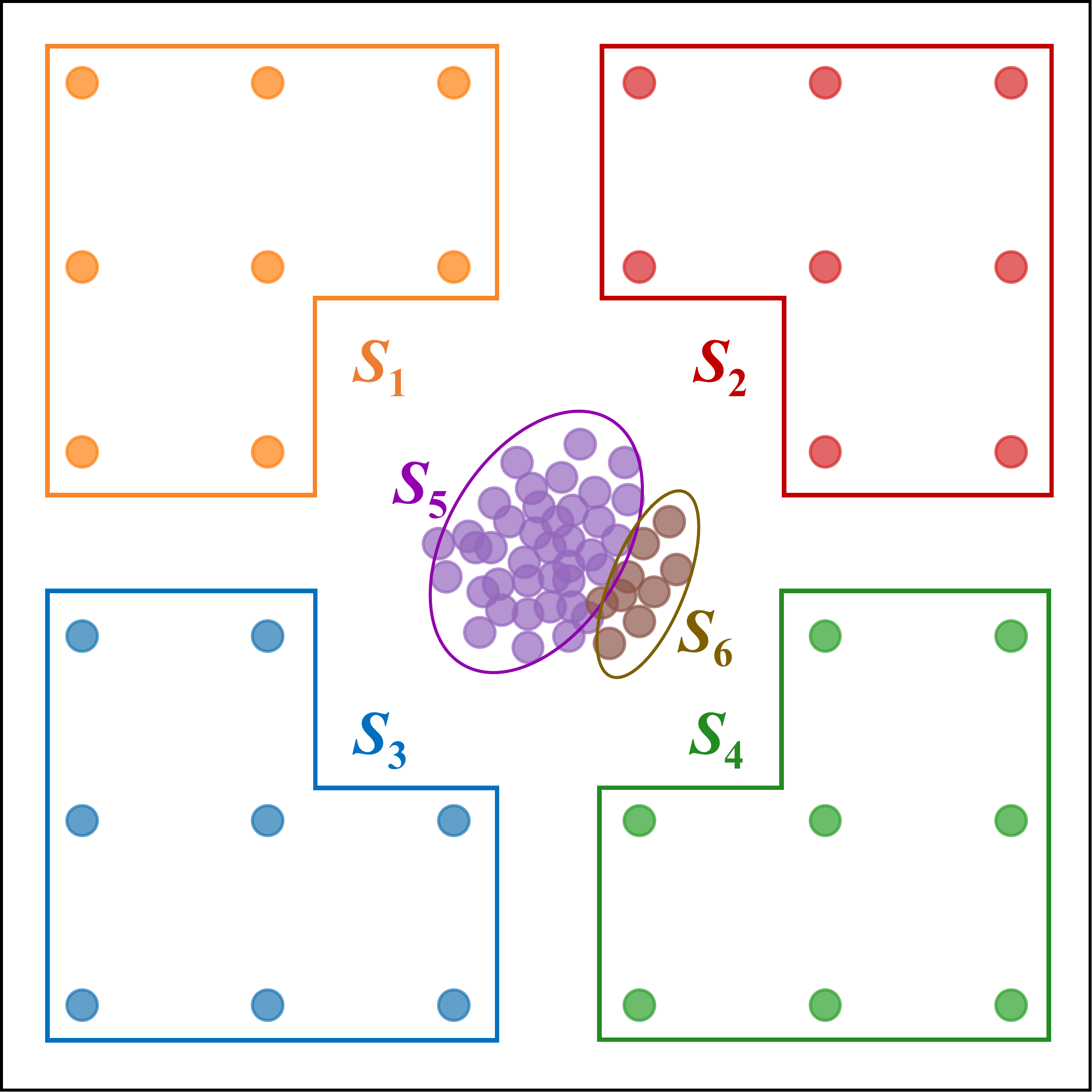}
}
\subfloat[]{
\label{Fig_DefectsInConnectivity3}
\includegraphics[width=0.29\columnwidth]{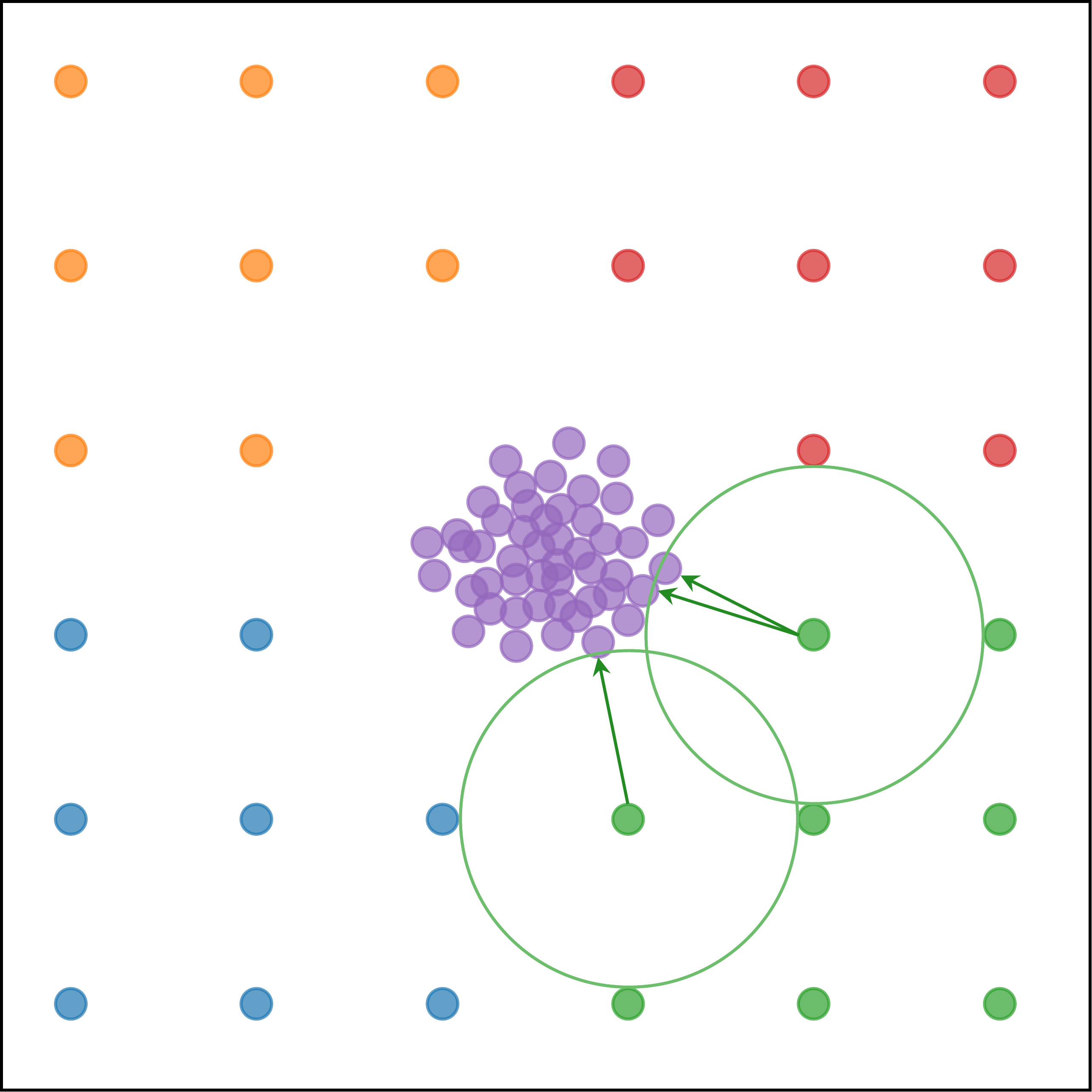}
}
\caption{The limitation of using only boundary connectivity for evaluating sub-cluster cohesion. (a) The ground truth of Pole\cite{CustomDatasets} dataset. (b) Initial partitioning of kFuse on the Pole. (c) Distance and adjacent samples between sparse cluster $S_4$ and dense cluster $S_6$.}
\label{Fig_DefectsInConnectivity}
\end{figure}

However, for datasets with significant density variations, $Con_{i,j}$ may not always be effective. Fig.~\ref{Fig_DefectsInConnectivity} demonstrates such a scenario. In Fig.~\ref{Fig_DefectsInConnectivity2}, sub-clusters $S_5$ and $S_6$ are very close and exhibit high boundary connectivity. While merging these two sub-clusters, the remaining sparse cluster may erroneously merge with the dense cluster due to their spatial proximity. For example, in Figure Fig.~\ref{Fig_DefectsInConnectivity3}, sub-cluster $S_4$ and the dense inner cluster $S_6$ have more adjacent samples and closer inter cluster distance, leading to higher boundary connectivity between them. To resolve this, we introduce density similarity.

\subsubsection{Density similarity}

For any sub-cluster $S_c,$ its density is defined as the mean density of all its samples, formally $\rho_c=\overline{\rho(o)}\ (o\in S_c)$. If $\sigma_c^2$ represents the variance in densities among the samples of $S_c$, then the density similarity $Sim\rho_{ij}$ between sub-clusters $S_i$ and $S_j$ is calculated using Equation~\ref{Eq_SimDensity}:

\begin{equation}
Sim\rho_{i,j}=\frac{\min(\rho_i,\rho_j)}{\max(\rho_i,\rho_j)} \times \left(1+\frac{\min(\sigma_i^2,\sigma_j^2)}{\max(\sigma_i^2,\sigma_j^2)}\right)
\label{Eq_SimDensity} 
\end{equation}
Where, the constant 1 is used to prevent sub-clusters with $\sigma_{c}^2=0$  from affecting the calculation.

From Equation~\ref{Eq_SimDensity}, we see that density similarity increases as the densities of the two sub-clusters become more similar, and the variance of sample densities within each sub-cluster decreases.

\begin{figure}[htbp]
\centering
\subfloat[Viewpoint 1]{
\label{Fig_SimilarityDensity1}
\includegraphics[width=0.45\columnwidth]{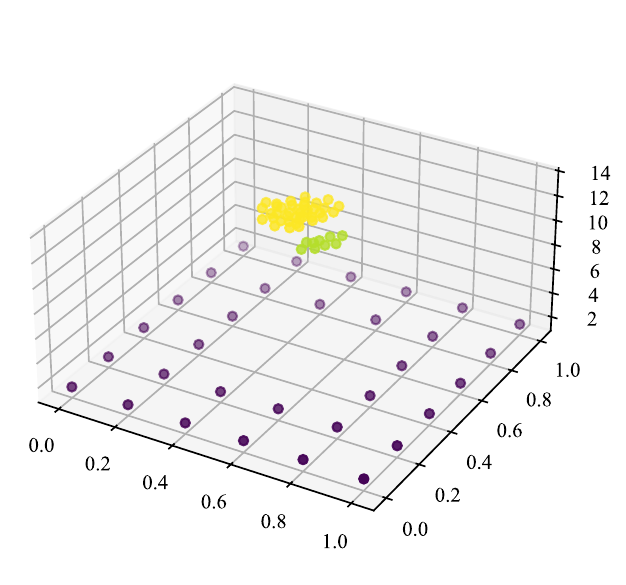}
}
\subfloat[Viewpoint 2]{
\label{Fig_SimilarityDensity2}
\includegraphics[width=0.45\columnwidth]{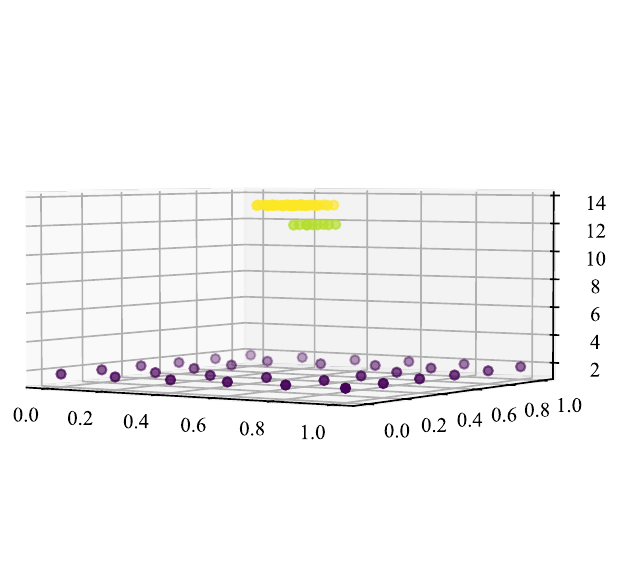}
}
\caption{Density differences between sub clusters in high-density datasets.}
\label{Fig_SimilarityDensity}
\end{figure}

Fig.~\ref{Fig_SimilarityDensity} illustrates the significant density differences between sparse and dense clusters. After introducing density similarity, sub-clusters with complex shapes and density variations can be better distinguished. It can be concluded that density similarity improves the accuracy of evaluating sub-cluster cohesion.

\subsubsection{Fusion index}

kFuse defines the fusion index $FI_{i,j}$ between sub-clusters $S_i$ and $S_j$ as the product of their boundary connectivity $Con_{i,j}$ and density similarity $Sim\rho_{i,j}$:

\begin{equation}
FI_{i,j}= Con_{i,j} \times Sim\rho_{i,j}
\label{Eq_FusionIndex} 
\end{equation}

The $FI_{i,j}\in(0,2)$ reflects the likelihood that sub-clusters $S_i$ and $S_j$ belong to the same cluster. Equation~\ref{Eq_FusionIndex} shows that $FI_{i,j}$ increases when $S_i$ and $S_j$ have more adjacent samples, are spatially closer, and have more similar densities.

After computing the fusion index for all sub-clusters, kFuse ranks them in descending order and stores the cluster pairs corresponding to each index in a queue. Finally, kFuse sequentially merges cluster pairs from this queue until the number of remaining sub-clusters equals the target number of clusters.

Algorithm~\ref{Alg_FuseClusters} provides a detailed explanation of the cluster merging process in kFuse, where $Mean(\cdot)$ calculates the mean of $\cdot$, and $Var(\cdot)$ calculates the variance of $\cdot$.

\begin{figure}[!t]
\begin{algorithm}[H]
\caption{Fused sub clusters}
\label{Alg_FuseClusters}
\begin{algorithmic}[1]
	\REQUIRE{$S, \rho, NN_\lambda, Labels, NC$ (number of clusters).}
	\ENSURE {$Labels$.} 
    \FOR{each sub cluster $S_c \in S$}
    {
        \STATE $\rho List_{S_c}=\{\rho(p)\mid p\in S_c\}$;
		\STATE $\rho_c = Mean(\rho List_{S_c})$;
		\STATE $\sigma_c^2=Var(\rho List_{S_c})$;
		\STATE $AS_c=S_c \cup \{\cup_{p\in S_c}NN_{\lambda}(p)\}$;
    }
    \ENDFOR
	\FOR{sub cluster $i \in \{ 1, 2, \dots, \lVert S\rVert - 1 \}$} 
	{
        \FOR{sub cluster $j \in \{ i+1, \dots, \lVert S\rVert \}$} 
    	{
            \STATE $Sim\rho_{i,j}=\frac{\min(\rho_i,\rho_j)}{\max(\rho_i,\rho_j)} \times \left(1+\frac{\min(\sigma_i^2,\sigma_j^2)}{\max(\sigma_i^2,\sigma_j^2)}\right)$;
			\STATE $ADS_{i,j}=AS_i\cap AS_j$;
			\STATE $DC_{i,j}= \min_{p\in S_i, q\in S_j}d(p,q)$
			\STATE $Con_{i,j}= \frac{(ADS_{i,j}+1)\times e^{(-DC_{i,j})}}{\min\left(\lVert S_i\rVert, \lVert S_j\rVert\right)}$;
			\STATE $FI_{i,j}= Sim\rho_{i,j} \times Con_{i,j}$;
        }
        \ENDFOR
    }
    \ENDFOR
    \STATE Sort the FI array in descending order and press the corresponding cluster pairs into the queue $Q$;
    \STATE $nc = \lVert S\rVert$;\\
	\WHILE{Q is not empty}
	{
        \STATE $(i, j)=Q.pop()$;
		\IF{$\exists (p \in S_i, q \in S_j) s.t. (Labels(p)\neq Labels(q)) $}
		{
			\STATE Fuse $S_1$ and $S_2$ (update $Labels$);
            \STATE $nc=nc - 1$;
		}
        \ENDIF
        \IF{$nc > NC$}
        {
            \STATE break;
        }
        \ENDIF
		\STATE $r=r+1$;
	}
    \ENDWHILE
    \RETURN $Labels$;
 \end{algorithmic}
\end{algorithm}	
\end{figure}

\subsection{Complexity Analysis}

Fig.~\ref{Fig_Process} presents the overall execution flow of kFuse, where the primary time complexity arises from three steps:

\begin{figure}
\centering
\includegraphics[width=0.95\columnwidth]{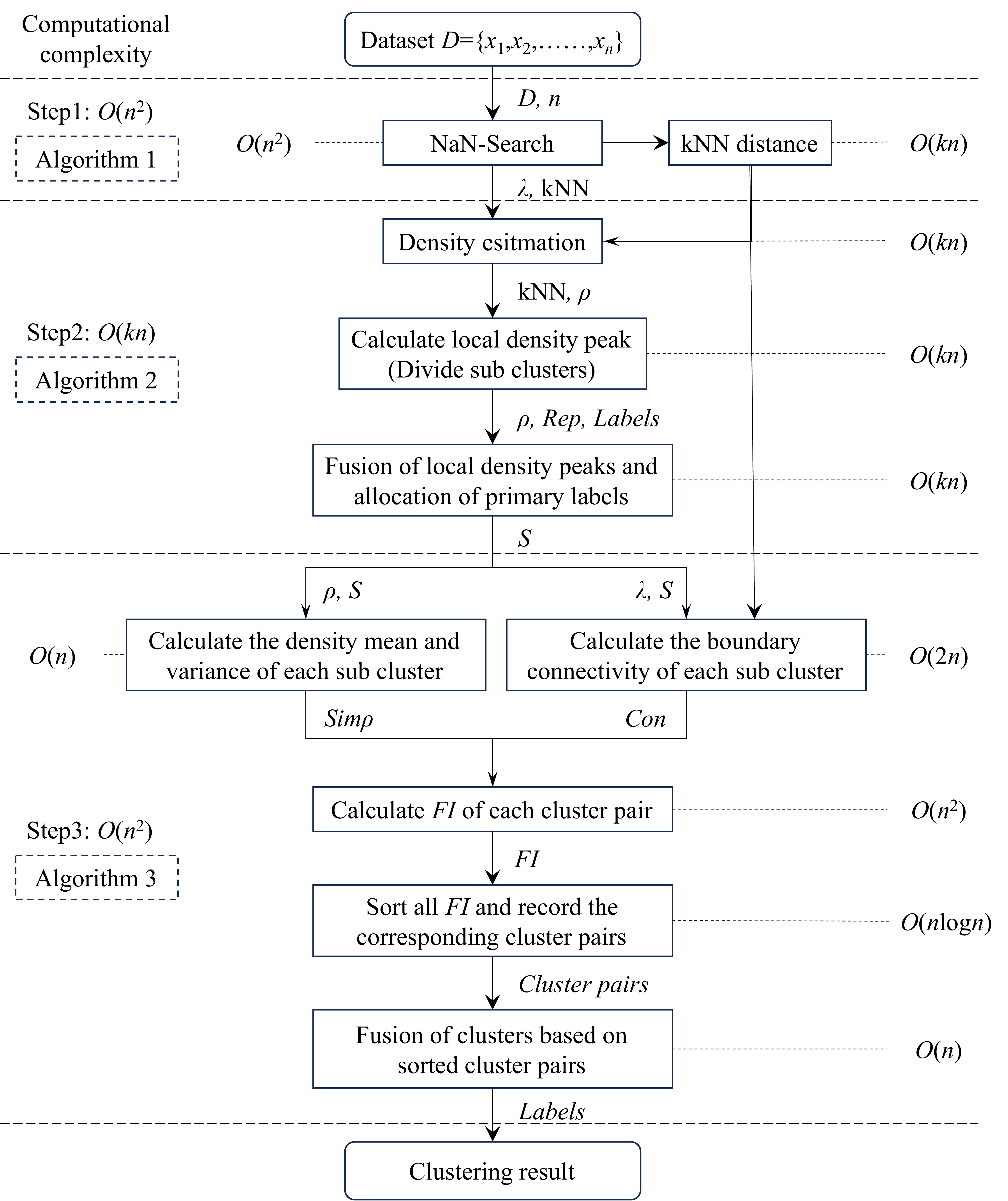}
\caption{The workflow of the overall kFuse Algorithm.}
\label{Fig_Process}
\end{figure}

\begin{enumerate}
\item{NaN-Searching(Algorithm~\ref{Alg_NaNSearch}). Obtaining the neighbor information of each sample, with time complexity $O(n^2)$ (where $n$ is the number of samples). Using $k = max(nb)$, the $k$-nearest neighbor distance matrix is obtained with a complexity of $O(kn)$, resulting in an overall complexity of $O(n^2)$.}
\item{Initial partitioning (Algorithm \ref{Alg_DividingClusters}). Calculating the local density and identifying local density peaks requires traversing each sample’s $k$-nearest neighbors, with a time complexity of $O(kn)$. Fusing local density peaks has a similar complexity of $O(kn)$. Thus, this step has a time complexity of $O(kn)$.}
\item{Cluster fusing (Algorithm \ref{Alg_FuseClusters}). Calculating boundary connectivity requires visiting all cluster pairs, with a worst-case complexity of $O\left({\frac{n}{k}}^2\right)$. Calculating the density and variance of each sub-cluster involves traversing all samples, with a time complexity of $O(n)$. The fusion index computation has a worst-case complexity of $O(\frac{n^2}{k^2})$. Sorting $FI$ values and recording cluster pairs has a worst-case time complexity of $O\left(\frac{n}{k}\log\left(\frac{n}{k}\right) \right)$. Finally, fusing sub-clusters requires visiting all samples, with the union-find data structure reducing the time complexity to $O(1)$, led to the time complexity of $O(n)$. Thus, this step has a time complexity of $O(n^2)$.}
\end{enumerate}

In summary, the overall time complexity of kFuse is $O(n^2 + kn + n^2)$, which simplifies to $O(n^2)$.

\section{Experiments}

\subsection{Experimental set up}

\textit{Datasets}. We selected 12 synthetic datasets with various shapes and 10 real-world datasets to evaluate the clustering performance of kFuse. Table~\ref{tab_Datasets} provides detailed information about these datasets.

\begin{table}[htbp]
\centering
\caption{Datasets}
\small
\setlength{\tabcolsep}{4pt}
\begin{tabular}{lcccc}
\toprule
Dataset & Instances & Attributes & Clusters & Source \\
\midrule
Pole & 80 & 2 & 2 & \cite{CustomDatasets} \\
Away & 229 & 2 & 3 & \cite{CustomDatasets} \\
Flame & 240 & 2 & 2 & \cite{ClusteringDatasets} \\
Zelnik1 & 299 & 2 & 3 & \cite{ZelnikDatasets} \\
Path-based & 300 & 2 & 3 & \cite{ClusteringDatasets} \\
Spiral & 312 & 2 & 3 & \cite{ClusteringDatasets} \\
Jain & 373 & 2 & 2 & \cite{ClusteringDatasets} \\
R15 & 600 & 2 & 15 & \cite{ClusteringDatasets} \\
Aggregation & 788 & 2 & 7 & \cite{ClusteringDatasets} \\
D31 & 3100 & 2 & 31 & \cite{ClusteringDatasets} \\
Unbalance & 6500 & 2 & 8 & \cite{ClusteringDatasets} \\
A3 & 7500 & 2 & 50 & \cite{ClusteringDatasets} \\
\midrule
Parkinson & 195 & 22 & 2 & \cite{UCIDatasets} \\
Ecoli & 336 & 7 & 8 & \cite{UCIDatasets} \\
LibrasMovement & 360 & 90 & 15 & \cite{UCIDatasets} \\
BreastCancer & 569 & 30 & 2 & \cite{UCIDatasets} \\
Yeast & 1484 & 8 & 10 & \cite{UCIDatasets} \\
Segmentation & 2310 & 19 & 7 & \cite{UCIDatasets} \\
Waveform & 5000 & 21 & 3 & \cite{UCIDatasets} \\
Musk & 6598 & 166 & 2 & \cite{UCIDatasets} \\
MNIST & 10000 & 784 & 10 & \cite{MnistDatasets}\\
OlivettiFaces & 400 & 2576 & 40 & \cite{OlivettiFaces}\\
\bottomrule
\end{tabular}
\label{tab_Datasets}
\end{table}

\textit{Comparative Algorithms and Settings}. The comparison includes the following algorithms: K-means\cite{Kmeans1966MJB}(a classic k-center clustering technique), AP\cite{AP2007BJF}(an effective non-parametric partitioning algorithm), DBSCAN\cite{DBSCAN1996EM} and MSC\cite{MeanShift} (density-based partitioning methods), DPC\cite{DPC2014AR} and ANN-DPC\cite{ANNDPC2024YH} (state-of-the-art density peak clustering algorithms), LDP-MST\cite{LDPMST2021CDD} (a clustering method also based on local density peaks), SLINK\cite{SLINK1973SR} (a classic single-link agglomerative clustering algorithm), and the proposed kFuse algorithm.

In terms of parameter settings, for K-means, ANN-DPC, LDP-MST, and kFuse, we used the correct number of clusters $C$ as input. For AP, DBSCAN, MSC, and SLINK, we selected the best parameter configuration from all possible settings. For DPC, we manually selected the correct number of clusters $C$ using an appropriate $d_c$ parameter and reported the best result. Additionally, for iterative algorithms like K-means, AP, and MSC, we selected the best result over 20 runs.

\textit{Data Preprocessing}. All datasets were preprocessed using min-max normalization to minimize the impact of different metrics across various dimensions\cite{Normalized2006PF}.

\textit{Evaluation Metrics}. We used the Fowlkes-Mallows Index (FMI)\cite{FMI1983EBF}, Adjusted Rand Index (ARI)\cite{ARI2009VNX}, and Normalized Mutual Information (NMI)\cite{NMI2003SA} to measure the clustering performance of each algorithm.

\subsection{Experiments on synthetic datasets}

In this section, we compare the proposed kFuse algorithm against other algorithms (LDP-MST\cite{LDPMST2021CDD}, SLINK\cite{SLINK1973SR}, DPC\cite{DPC2014AR}, ANN-DPC\cite{ANNDPC2024YH}, MSC\cite{MeanShift}, AP\cite{AP2007BJF}, K-means\cite{Kmeans1966MJB}, DBSCAN\cite{DBSCAN1996EM}) on 12 synthetic datasets of different shapes.

\begin{figure*}
\centering
\includegraphics[width=0.98\textwidth]{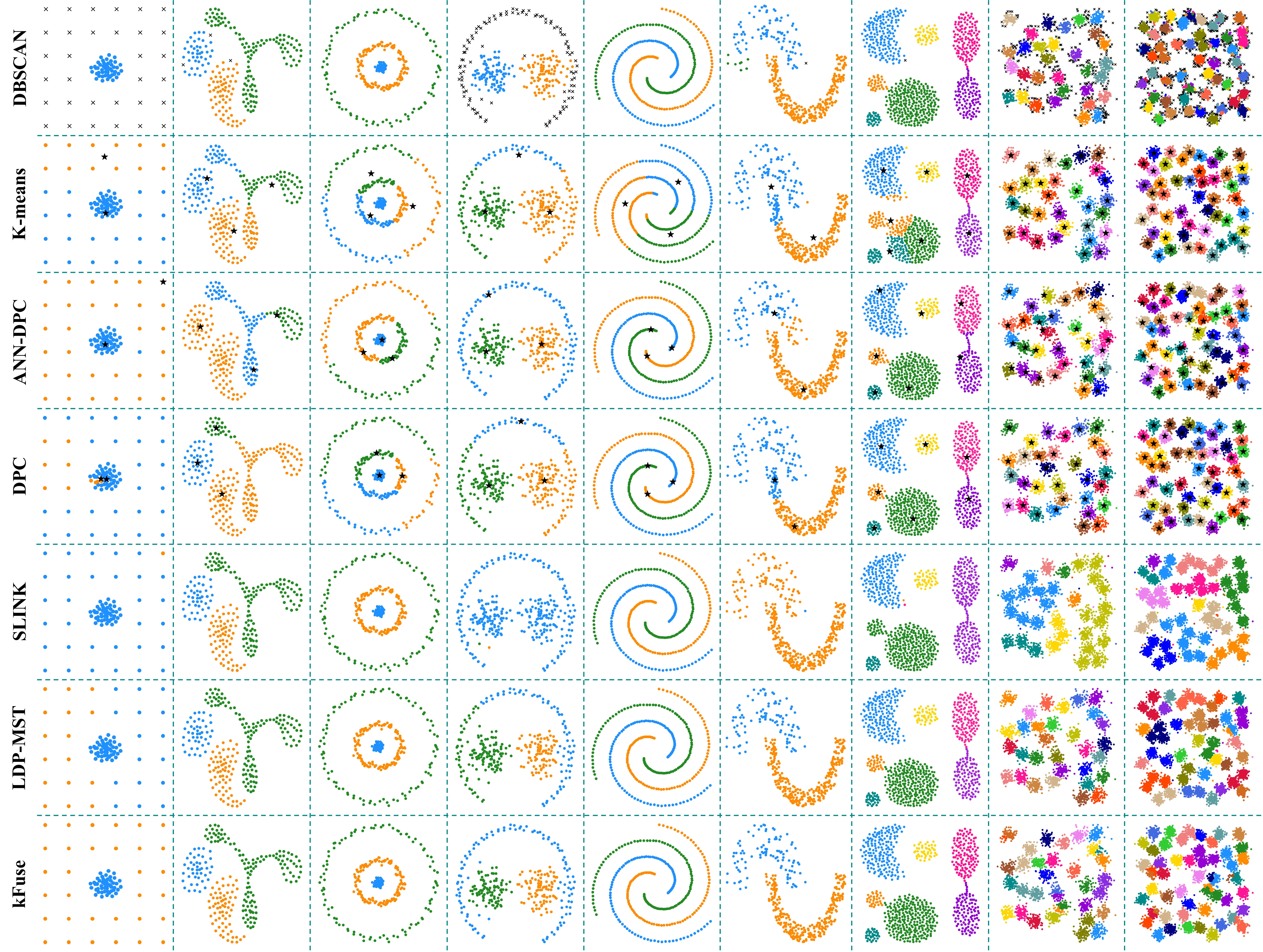}
\caption{The results of different algorithms on synthetic datasets. The datasets from left to right are: Pole, Away, Zelnik1, Path-based, Spiral, Jain, Aggregation, D31, A3.}
\label{Fig_SYNtest}
\end{figure*}

Fig.~\ref{Fig_SYNtest} shows the comparison results between different algorithms, where "$\star$" represents the cluster centers identified by K-means, DPC, and ANN-DPC, and "$\times$" denotes noise points identified by DBSCAN. As shown, kFuse, using the local density peak-based merging strategy, almost perfectly identifies all clusters across the datasets. LDP-MST demonstrates excellent potential in recognizing ring-shaped and spiral clusters but fails to differentiate between the two clusters of different densities in the high-density variation dataset "Pole." Moreover, LDP-MST struggles with path-based clusters, which have fuzzy boundaries. SLINK performs well on datasets with clear boundaries, such as "Away," "Zelnik1," and "Spiral," but it fails to identify clusters in datasets with large density variations like "Pole," as well as fuzzy-boundary datasets like "Flame," "Path-based," "Aggregation," "D31," and "A3." DPC achieves satisfactory results on "Flame," "Spiral," "Aggregation," "D31," and "A3," but fails on datasets like "Pole," "Away," "Zelnik1," "Path-based," and "Jain" due to incorrect cluster center identification. ANN-DPC, a recent density peak clustering algorithm that can automatically select cluster centers, correctly identifies good cluster centers on datasets like "Pole," "Flame," "Path-based," "Spiral," "Jain," and "Aggregation," but fails on more complex-shaped datasets like "Away" and "Zelnik1." K-means cannot identify non-spherical clusters, and DBSCAN nearly reconstructs all shapes but misidentifies many boundary points in "Pole," "Path-based," "D31," and "A3" as noise.

To further assess performance, Table~\ref{tab_ResultOnSYN} presents detailed FMI, ARI, and NMI scores for each algorithm across various datasets, with the best results highlighted in bold. As shown, the proposed kFuse achieves high scores on most datasets (8/12), with LDP-MST as a close competitor. ANN-DPC outperforms DPC on five datasets, while AP and K-means lose their competitiveness in recognizing non-spherical clusters. MSC, another density-based method, performs overall worse than DBSCAN.

\begin{table*}
\centering	
\caption{The Comparison of FMI, ARI and NMI on Synthetic Datasets}
\small
\setlength{\tabcolsep}{4pt}
\begin{tabular}{lccccccccccc}
\toprule
Datasets & Metrics & kFuse & LDP-MST & SLINK & ANN-DPC & DPC & MSC & AP & K-means & DBSCAN\\
\midrule
\multirow{3}{*}{Pole}
& \emph{FMI} & \textbf{1} & 0.734 & 0.712 & 0.975 & 0.634 & 0.873 & 0.641 & 0.709 & \textbf{1} \\
& \emph{ARI} & \textbf{1} & 0.345 & 0.012 & 0.949 & 0.024 & 0.758 & 0.404 & 0.230 & \textbf{1} \\
& \emph{NMI} & \textbf{1} & 0.380 & 0.031 & 0.913 & 0.012 & 0.708 & 0.581 & 0.288 & \textbf{1} \\
\midrule
\multirow{3}{*}{Away} 
& \emph{FMI} & \textbf{1} & \textbf{1} & \textbf{1} & 0.720 & 0.650 & 0.614 & 0.504 & 0.565 & 0.992 \\
& \emph{ARI} & \textbf{1} & \textbf{1} & \textbf{1} & 0.541 & 0.353 & 0.431 & 0.296 & 0.313 & 0.987 \\
& \emph{NMI} & \textbf{1} & \textbf{1} & \textbf{1} & 0.688 & 0.594 & 0.621 & 0.605 & 0.451 & 0.968 \\
\midrule
\multirow{3}{*}{Flame}
& \emph{FMI} & \textbf{1} & 0.969 & 0.730 & 0.984 & \textbf{1} & 0.764 & 0.468 & 0.736 & 0.976 \\
& \emph{ARI} & \textbf{1} & 0.933 & 0.012 & 0.966 & \textbf{1} & 0.511 & 0.204 & 0.453 & 0.949 \\
& \emph{NMI} & \textbf{1} & 0.889 & 0.024 & 0.926 & \textbf{1} & 0.458 & 0.422 & 0.398 & 0.890 \\
\midrule
\multirow{3}{*}{Zelnik1}
& \emph{FMI} & \textbf{1} & \textbf{1} & \textbf{1} & 0.566 & 0.415 & 0.662 & 0.487 & 0.402 & \textbf{1} \\
& \emph{ARI} & \textbf{1} & \textbf{1} & \textbf{1} & 0.320 & 0.080 & 0.423 & 0.283 & 0.050 & \textbf{1} \\
& \emph{NMI} & \textbf{1} & \textbf{1} & \textbf{1} & 0.485 & 0.181 & 0.549 & 0.569 & 0.160 & \textbf{1} \\
\midrule
\multirow{3}{*}{Path-based}
& \emph{FMI} & \textbf{0.972} & 0.782 & 0.573 & 0.959 & 0.666 & 0.708 & 0.528 & 0.661 & 0.934 \\
& \emph{ARI} & \textbf{0.959} & 0.668 & 0.001 & 0.939 & 0.471 & 0.581 & 0.345 & 0.461 & 0.901 \\
& \emph{NMI} & \textbf{0.941} & 0.747 & 0.013 & 0.911 & 0.554 & 0.586 & 0.576 & 0.545 & 0.872 \\
\midrule
\multirow{3}{*}{Spiral}
& \emph{FMI} & \textbf{1} & \textbf{1} & \textbf{1} & \textbf{1} & \textbf{1} & 0.309 & 0.332 & 0.327 & \textbf{1} \\
& \emph{ARI} & \textbf{1} & \textbf{1} & \textbf{1} & \textbf{1} & \textbf{1} & 0.124 & 0.151 & -0.006 & \textbf{1} \\
& \emph{NMI} & \textbf{1} & \textbf{1} & \textbf{1} & \textbf{1} & \textbf{1} & 0.471 & 0.434 & 0.001 & \textbf{1} \\
\midrule
\multirow{3}{*}{Jain}
& \emph{FMI} & \textbf{1} & \textbf{1} & 0.784 & 0.915 & 0.816 & 0.816 & 0.392 & 0.820 & 0.990 \\
& \emph{ARI} & \textbf{1} & \textbf{1} & 0.009 & 0.757 & 0.569 & 0.569 & 0.123 & 0.576 & 0.975 \\
& \emph{NMI} & \textbf{1} & \textbf{1} & 0.012 & 0.674 & 0.542 & 0.542 & 0.380 & 0.527 & 0.928 \\
\midrule
\multirow{3}{*}{R15}
& \emph{FMI} & 0.986 & 0.986 & 0.637 & 0.983 & \textbf{0.993} & \textbf{0.993} & \textbf{0.993} & \textbf{0.993} & 0.983 \\
& \emph{ARI} & 0.985 & 0.985 & 0.542 & 0.982 & \textbf{0.992} & \textbf{0.992} & \textbf{0.992} & \textbf{0.992} & 0.981 \\
& \emph{NMI} & 0.989 & 0.989 & 0.877 & 0.986 & \textbf{0.994} & \textbf{0.994} & \textbf{0.994} & \textbf{0.994} & 0.986 \\
\midrule
\multirow{3}{*}{Aggregation}
& \emph{FMI} & 0.994 & \textbf{0.996} & 0.862 & 0.977 & \textbf{0.996} & 0.878 & 0.556 & 0.776 & 0.990 \\
& \emph{ARI} & 0.993 & \textbf{0.995} & 0.805 & 0.971 & \textbf{0.995} & 0.831 & 0.413 & 0.715 & 0.988 \\
& \emph{NMI} & 0.989 & \textbf{0.992} & 0.884 & 0.965 & \textbf{0.992} & 0.892 & 0.763 & 0.829 & 0.982 \\
\midrule
\multirow{3}{*}{D31}
& \emph{FMI} & 0.822 & 0.930 & 0.349 & 0.889 & 0.938 & 0.937 & 0.955 & \textbf{0.956} & 0.711 \\
& \emph{ARI} & 0.811 & 0.928 & 0.173 & 0.886 & 0.936 & 0.935 & \textbf{0.954} & \textbf{0.954} & 0.697 \\
& \emph{NMI} & 0.928 & 0.953 & 0.637 & 0.938 & 0.957 & 0.957 & \textbf{0.968} & \textbf{0.968} & 0.872 \\
\midrule
\multirow{3}{*}{Unbalance}
& \emph{FMI} & \textbf{1} & \textbf{1} & \textbf{1} & 0.742 & \textbf{1} & 0.999 & 0.496 & \textbf{1} & 0.999 \\
& \emph{ARI} & \textbf{1} & \textbf{1} & \textbf{1} & 0.645 & \textbf{1} & 0.999 & 0.318 & \textbf{1} & 0.999 \\
& \emph{NMI} & \textbf{1} & \textbf{1} & \textbf{1} & 0.736 & \textbf{1} & 0.998 & 0.660 & \textbf{1} & 0.998 \\
\midrule
\multirow{3}{*}{A3}
& \emph{FMI} & 0.839 & 0.712 & 0.448 & 0.882 & 0.982 & 0.995 & 0.994 & \textbf{0.996} & 0.708 \\
& \emph{ARI} & 0.831 & 0.668 & 0.315 & 0.879 & 0.982 & 0.995 & 0.994 & \textbf{0.996} & 0.697 \\
& \emph{NMI} & 0.956 & 0.917 & 0.761 & 0.955 & 0.988 & 0.996 & 0.996 & \textbf{0.997} & 0.887 \\
\midrule
\multirow{3}{*}{Average}
& \emph{FMI} & \textbf{0.968} & 0.926 & 0.758 & 0.883 & 0.841 & 0.796 & 0.612 & 0.745 & 0.940 \\
& \emph{ARI} & \textbf{0.965} & 0.877 & 0.488 & 0.820 & 0.700 & 0.679 & 0.456 & 0.561 & 0.931 \\
& \emph{NMI} & \textbf{0.984} & 0.906 & 0.603 & 0.848 & 0.735 & 0.731 & 0.662 & 0.597 & 0.949 \\
\bottomrule
\end{tabular} 
\label{tab_ResultOnSYN}
\end{table*}

The final row of Table~\ref{tab_ResultOnSYN} summarizes the average FMI, ARI, and NMI scores across all datasets for each algorithm. Our kFuse achieves the best average scores in all three metrics, demonstrating its superior clustering performance on synthetic datasets.

\subsection{Experiments on real-world datasets}

 Real-world datasets tend to be high-dimensional and large-scale, posing significant challenges for clustering algorithms. This makes their performance in practical applications particularly important.

In this subsection, we evaluate the performance of the algorithms on 10 real-world datasets, including 8 UCI datasets\cite{UCIDatasets} (Parkinson, Ecoli, LibrasMovement, BreastCancer, Yeast, Segmentation, Waveform, and Musk) and 2 popular large-scale machine learning datasets (MNIST\cite{MnistDatasets} and OlivettiFaces\cite{OlivettiFaces}). Table~\ref{tab_ResultOnReal} presents the experimental results, with the best scores highlighted in bold.

\begin{table*}
\centering	
\caption{The Comparison of FMI, ARI and NMI on Real-World Datasets}
\small
\setlength{\tabcolsep}{4pt}
\begin{tabular}{lccccccccccc}
\toprule
Datasets & Metrics & kFuse & LDP-MST & SLINK & ANN-DPC & DPC & MSC & AP & K-means & DBSCAN\\
\midrule
\multirow{3}{*}{Parkinson} 
& \emph{FMI} & \textbf{0.819} & 0.704 & 0.607 & 0.810 & 0.742 & 0.402 & 0.640 & 0.594 & 0.813\\
& \emph{ARI} & \textbf{0.363} & 0.296 & -0.083 & 0.276 & -0.051 & 0.077 & 0.174 & 0.045 & 0.282 \\
& \emph{NMI} & \textbf{0.271} & 0.211 & 0.135 & 0.221 & 0.036 & 0.184 & 0.143 & 0.231 & 0.245 \\
\midrule
\multirow{3}{*}{Ecoli}
& \emph{FMI} & 0.631 & 0.518 & 0.538 & \textbf{0.692} & 0.576 & 0.613 & 0.414 & 0.577 & 0.587 \\
& \emph{ARI} & 0.510 & 0.373 & 0.399 & \textbf{0.595} & 0.239 & 0.489 & 0.251 & 0.446 & 0.457 \\
& \emph{NMI} & 0.595 & 0.371 & 0.603 & \textbf{0.654} & 0.378 & 0.436 & 0.563 & 0.598 & 0.416 \\
\midrule
\multirow{3}{*}{LibrasMovement}
& \emph{FMI} & 0.303 & 0.297 & 0.371 & \textbf{0.477} & 0.285 & 0.318 & 0.341 & 0.356 & 0.248 \\
& \emph{ARI} & 0.110 & 0.105 & 0.319 & \textbf{0.373} & 0.108 & 0.264 & 0.297 & 0.306 & 0.003 \\
& \emph{NMI} & 0.393 & 0.295 & 0.615 & 0.628 & 0.491 & \textbf{0.686} & 0.650 & 0.600 & 0.080 \\
\midrule
\multirow{3}{*}{BreastCancer}
& \emph{FMI} & \textbf{0.879} & 0.829 & 0.788 & 0.820 & 0.728 & 0.773 & 0.644 & 0.877 & 0.581 \\
& \emph{ARI} & \textbf{0.736} & 0.637 & 0.538 & 0.627 & 0.002 & 0.567 & 0.352 & 0.730 & 0.199 \\
& \emph{NMI} & \textbf{0.626} & 0.514 & 0.417 & 0.545 & 0.005 & 0.436 & 0.377 & 0.623 & 0.238\\
\midrule
\multirow{3}{*}{Yeast}
& \emph{FMI} & \textbf{0.456} & 0.426 & 0.298 & 0.319 & 0.447 & 0.326 & 0.126 & 0.310 & 0.370 \\
& \emph{ARI} & 0.129 & \textbf{0.295} & 0.144 & 0.139 & 0.060 & 0.143 & 0.034 & 0.159 & 0.053 \\
& \emph{NMI} & 0.216 & 0.286 & 0.267 & 0.225 & 0.114 & 0.266 & 0.263 & \textbf{0.287} & 0.100 \\
\midrule
\multirow{3}{*}{Segmentation}
& \emph{FMI} & \textbf{0.686} & 0.504 & 0.588 & 0.516 & 0.430 & 0.580 & 0.337 & 0.575 & 0.514 \\
& \emph{ARI} & \textbf{0.552} & 0.422 & 0.518 & 0.409 & 0.100 & 0.515 & 0.200 & 0.504 & 0.440 \\
& \emph{NMI} & \textbf{0.659} & 0.415 & 0.640 & 0.612 & 0.344 & 0.615 & 0.584 & 0.612 & 0.633 \\
\midrule
\multirow{3}{*}{Waveform}
& \emph{FMI} & 0.547 & 0.574 & 0.586 & 0.532 & 0.561 & \textbf{0.633} & 0.106 & 0.503 & 0.576 \\
& \emph{ARI} & 0.034 & 0.362 & \textbf{0.379} & 0.290 & 0.002 & 0.367 & 0.016 & 0.253 & 0.000 \\
& \emph{NMI} & 0.141 & 0.306 & \textbf{0.415} & 0.376 & 0.019 & 0.378 & 0.234 & 0.364 & 0.001 \\
\midrule
\multirow{3}{*}{Musk}
& \emph{FMI} & \textbf{0.851} & 0.833 & 0.619 & 0.773 & 0.609 & 0.044 & 0.074 & 0.618 & 0.737 \\
& \emph{ARI} & 0.041 & 0.002 & -0.033 & -0.055 & 0.001 & 0.000 & 0.002 & -0.032 & \textbf{0.126} \\
& \emph{NMI} & 0.065 & 0.004 & 0.027 & 0.011 & 0.001 & 0.097 & 0.093 & 0.028 & \textbf{0.128} \\
\midrule
\multirow{3}{*}{MNIST}
& \emph{FMI} & \textbf{0.852} & 0.701 & 0.316 & 0.583 & 0.715 & 0.689 & 0.357 & 0.763 & 0.396 \\
& \emph{ARI} & \textbf{0.836} & 0.564 & 0.174 & 0.428 & 0.619 & 0.524 & 0.082 & 0.723 & 0.240 \\
& \emph{NMI} & \textbf{0.896} & 0.672 & 0.228 & 0.491 & 0.676 & 0.599 & 0.482 & 0.801 & 0.564 \\
\midrule
\multirow{3}{*}{OlivettiFaces}
& \emph{FMI} & \textbf{0.272} & 0.209 & 0.158 & 0.187 & 0.223 & 0.245 & 0.194 & 0.228 & 0.172 \\
& \emph{ARI} & 0.160 & 0.058 & 0.008 & 0.168 & 0.118 & \textbf{0.202} & 0.138 & 0.162 & 0.103 \\
& \emph{NMI} & \textbf{0.758} & 0.487 & 0.220 & 0.552 & 0.616 & 0.519 & 0.425 & 0.669 & 0.421 \\
\midrule
\multirow{3}{*}{Average}
& \emph{FMI} & \textbf{0.630} & 0.559 & 0.487 & 0.571 & 0.532 & 0.462 & 0.323 & 0.540 & 0.499 \\
& \emph{ARI} & \textbf{0.347} & 0.311 & 0.236 & 0.325 & 0.120 & 0.315 & 0.155 & 0.330 & 0.190 \\
& \emph{NMI} & 0.462 & 0.356 & 0.357 & 0.432 & 0.268 & 0.422 & 0.381 & \textbf{0.481} & 0.283 \\
\bottomrule
\end{tabular} 
\label{tab_ResultOnReal}
\end{table*}

As shown in Table~\ref{tab_ResultOnReal}, kFuse achieves the best FMI scores on 7/10 real-world datasets, the best ARI scores on 4/10 datasets, and the best NMI scores on 5/10 datasets. The final row of Table~\ref{tab_ResultOnSYN} summarizes the average scores across all datasets, showing that kFuse achieves the best average scores in FMI and ARI. The best average NMI score of 0.481 is obtained by K-means, suggesting that K-means still has some advantages on real-world datasets. kFuse's average NMI score of 0.462 ranks second with a narrow margin.

Overall, kFuse demonstrates outstanding performance, indicating its effectiveness in recognizing clusters in real-world datasets.

\subsection{The handwritten digit recognition of MNIST}

In handwritten digit recognition, each class often consists of multiple sub-classes, as different users write the same digit in different ways\cite{DataClustering2010JA}, posing a significant challenge for many clustering algorithms. We used a target dataset of 10,000 test samples (each sample is 28×28 pixels) from the \cite{MnistDatasets} to evaluate the performance of our proposed algorithm.

\begin{figure}[htbp]
\centering
\includegraphics[width=0.96\columnwidth]{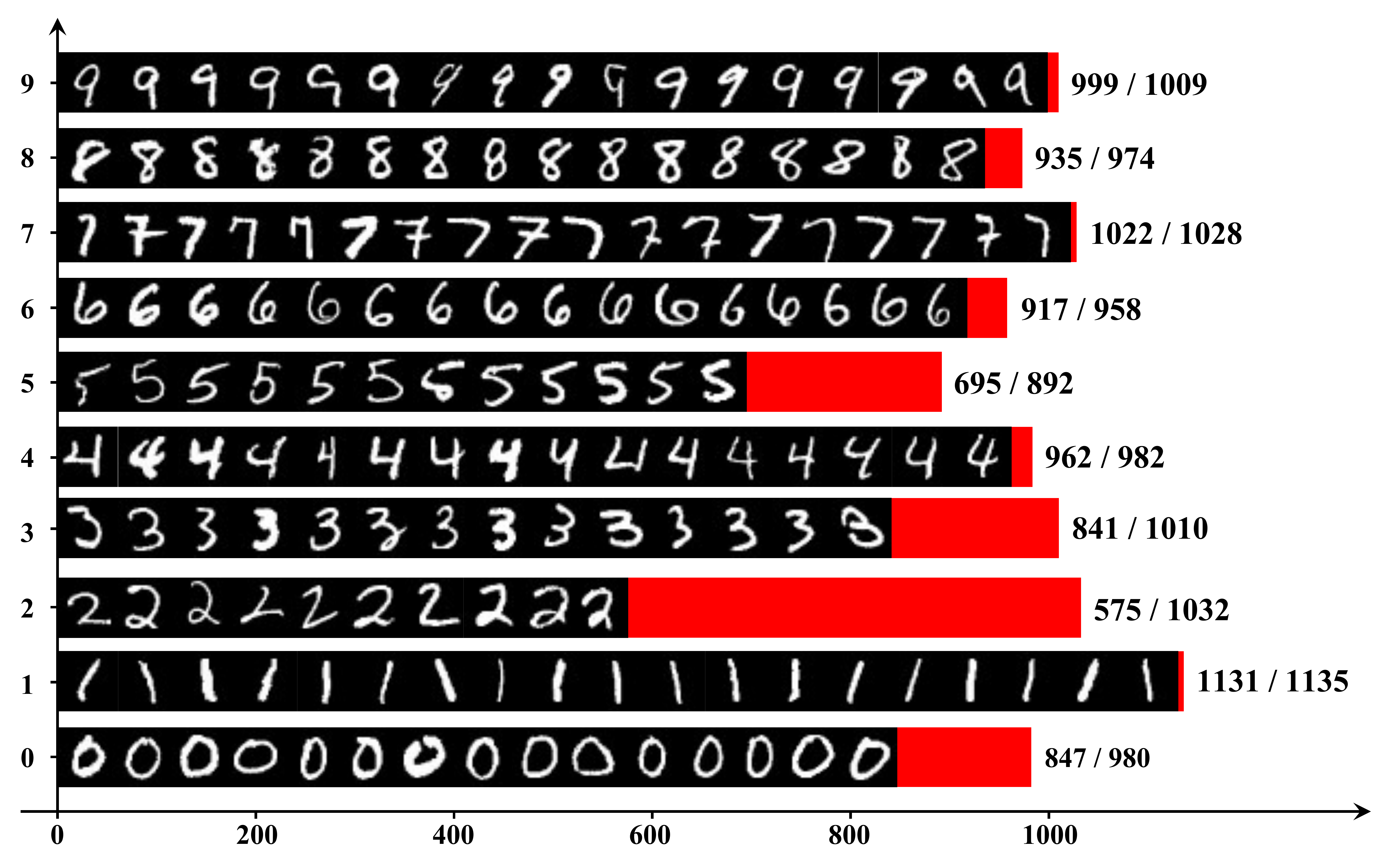}
\caption{KFuse's recognition details on MNIST.}
\label{Fig_MNIST}
\end{figure}

Table~\ref{tab_ResultOnReal} presents the clustering evaluation results for each algorithm on the MNIST dataset. As seen, kFuse achieves the highest FMI, NMI, and ARI scores on MNIST, followed by K-means and DPC. Fig.~\ref{Fig_MNIST} illustrates the details of kFuse's recognition of different digits in the MNIST dataset.

From Fig.~\ref{Fig_MNIST}, we can see that kFuse performs slightly less effectively in recognizing digits "2," "3," and "5," but it accurately identifies the other digits, demonstrating its effectiveness in handwritten digit recognition.

\subsection{The face recognition of Olivetti Faces}

The Olivetti Faces\cite{OlivettiFaces} contains 400 face images of 40 individuals, with 10 samples per individual under different poses, expressions, and lighting conditions. Accurately obtaining 40 face clusters is quite challenging due to the relatively high cluster count compared to the total number of data points\cite{DPC2014AR}.

Table~\ref{tab_ResultOnReal} presents the clustering evaluation results for each algorithm on the Olivetti Faces dataset. kFuse achieves the highest FMI and NMI scores on Olivetti Faces. In terms of ARI, MSC attains the highest score, followed by ANN-DPC and K-means, while kFuse ranks fourth. Nevertheless, kFuse's NMI score is 0.160, only 0.042 behind the top ranked MSC's score of 0.202.

\begin{figure}[htbp]
\centering
\includegraphics[width=0.96\columnwidth]{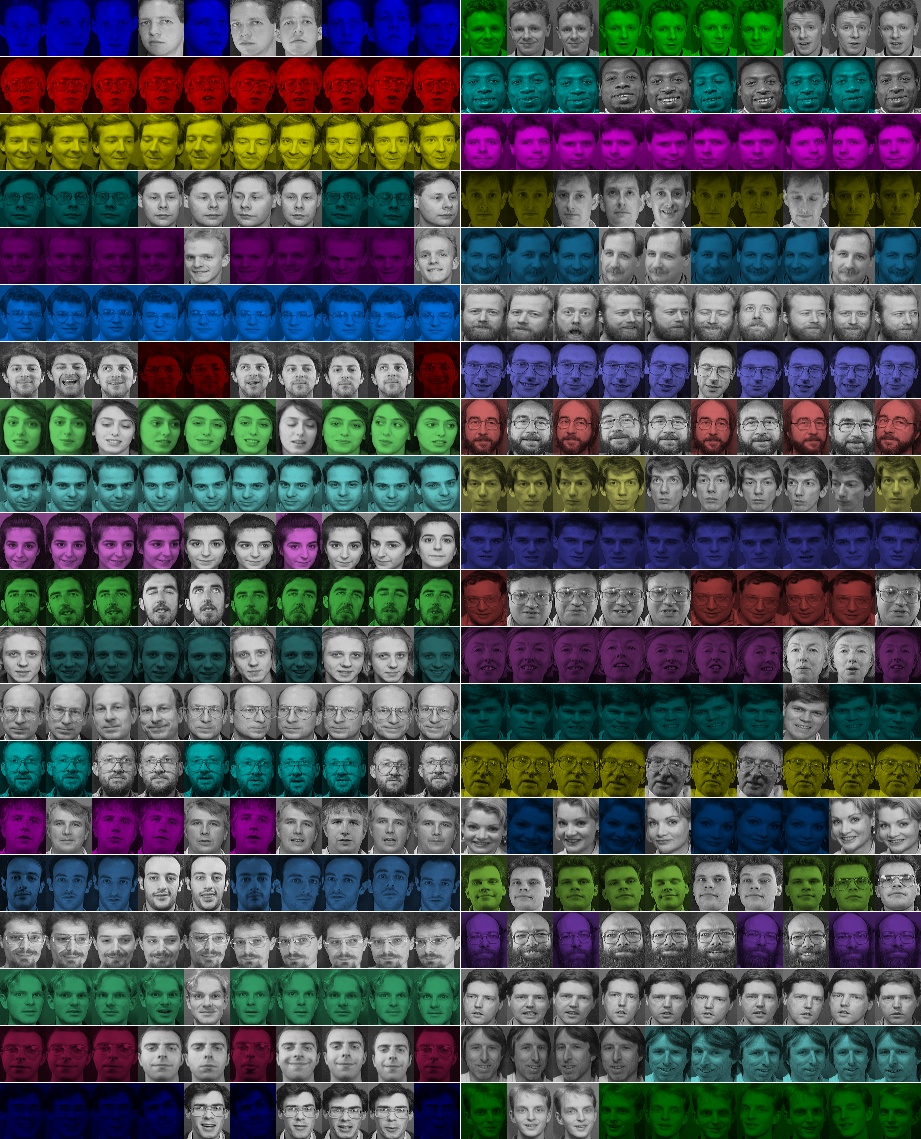}
\caption{KFuse's recognition details on Olivetti Faces.}
\label{Fig_Olivetti}
\end{figure}

Fig.~\ref{Fig_Olivetti} illustrates kFuse's recognition of different faces in the Olivetti Faces dataset, where gray faces represent unidentified ones, and faces with the same filter represent those clustered into the same group. Fig.~\ref{Fig_Olivetti} shows that kFuse identified 36/40 faces, with 5 of them perfectly clustered. This demonstrates that kFuse is effective in face recognition.

The above experiments demonstrate that the proposed kFuse algorithm can correctly and effectively cluster both synthetic and real-world datasets.

\subsection{Statistical Significance Tests}

This section performs statistical tests on the 9 clustering algorithms to verify whether the performance differences between kFuse and other algorithms (LDP-MST, SLINK, DPC, ANN-DPC, MSC, AP, K-means, and DBSCAN) are statistically significant.

First, the Friedman test is applied to detect significant differences among all 9 algorithms. Once significant differences are detected, the Nemenyi post-hoc test is conducted to determine pairwise significance. The significance level for the Friedman test is set at $\alpha=0.05$. This test evaluates the FMI, ARI, and NMI metrics of the 9 algorithms across 12 synthetic datasets, 8 UCI real-world datasets, and the MNIST and Olivetti Faces datasets, for a total of 22 datasets. The results of the Friedman test are presented in Table~\ref{tab_Friedman}.

\begin{table}
\centering	
\caption{Friedman test results}
\begin{tabular}{cccc}
\toprule
Result & FMI & ARI & NMI\\
\midrule
$\chi^2$ & 36.80 & 26.19 & 17.69 \\
$df$ & 8 & 8 & 8 \\
$p$ & 1.2506e-05 & 9.7521e-04 & 2.3713e-02 \\
\bottomrule
\end{tabular} 
\label{tab_Friedman}
\end{table}

The results in Table~\ref{tab_Friedman} show that the $p$-values for the Friedman test on FMI, ARI, and NMI are all less than 0.05, indicating significant differences among the algorithms. Therefore, the Nemenyi post-hoc test is necessary to assess the statistical significance of pairwise algorithm comparisons.

\begin{figure*}[htbp]
\centering
\subfloat[FMI]{
\label{Fig_NemenyiFMI}
\includegraphics[width=0.31\textwidth]{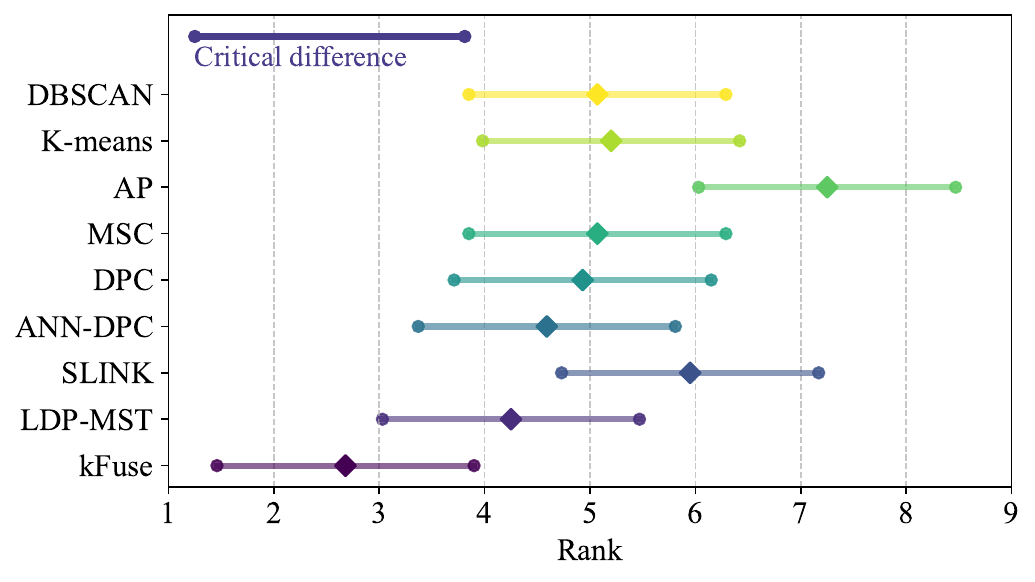}
}
\subfloat[ARI]{
\label{Fig_NemenyiARI}
\includegraphics[width=0.31\textwidth]{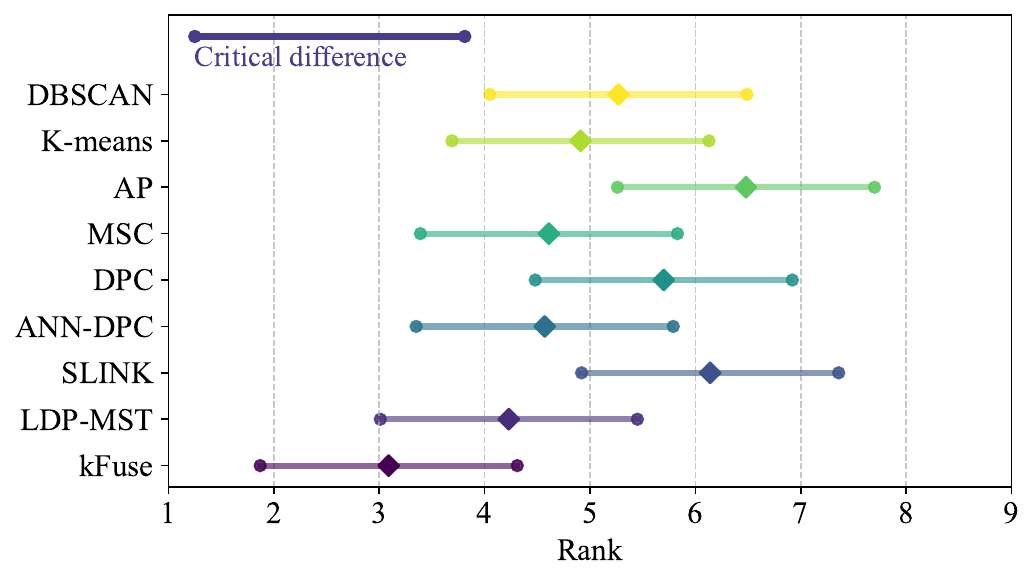}
}
\subfloat[NMI]{
\label{Fig_NemenyiNMI}
\includegraphics[width=0.31\textwidth]{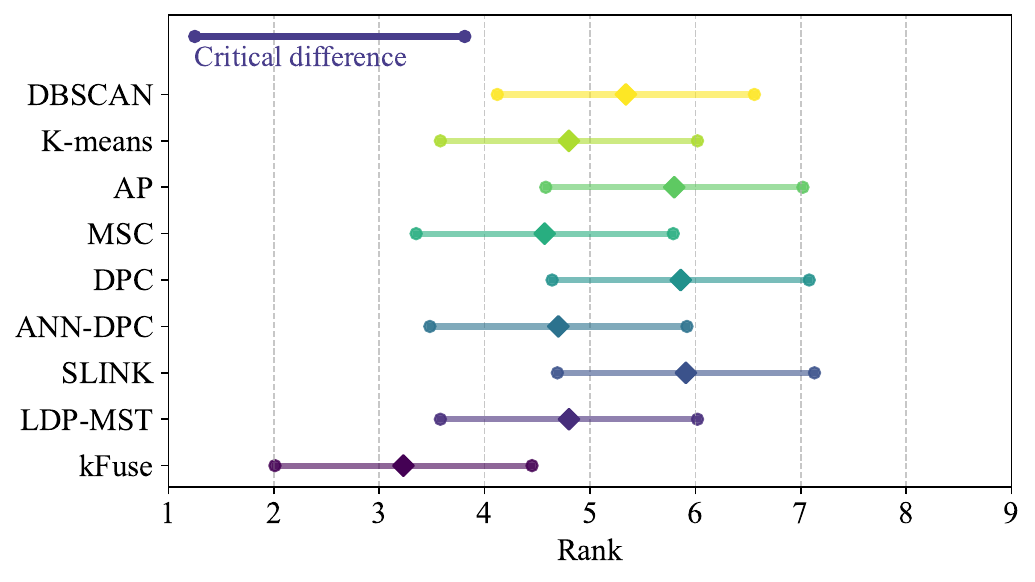}
}
\caption{Comparison of nine algorithms against each other utilizing Nemenyi’s test.}
\label{Fig_Nemenyi}
\end{figure*}

The Nemenyi test calculates the average rank and the critical difference ($CD$) between algorithms. If the average rank difference between two algorithms is less than $CD$, we accept the null hypothesis that there is no significant difference between the algorithms at a confidence level of $1-\alpha$. Otherwise, we reject this hypothesis, indicating a statistically significant difference between the algorithms. The formula for calculating $CD$ is as follows:

\begin{equation}
CD=q_\alpha\sqrt{\frac{M(M+1)}{6N}}
\label{Eq_CD} 
\end{equation}
Where $M$ represents the number of algorithms being compared, $N$ represents the number of datasets in the experiment, and $q_\alpha$ is a value obtained from the table based on $M$ and $\alpha$. In this experiment, $M = 9, N = 22$, and $q_\alpha = 3.102$, yielding a $CD$ value of 2.5614. The results of the Nemenyi post-hoc test at a 0.95 confidence level are shown in Fig.~\ref{Fig_Nemenyi}.

As shown in Fig.~\ref{Fig_Nemenyi}, kFuse has statistically significant differences in FMI compared to SLINK and AP. While kFuse does not exhibit significant differences in FMI with other algorithms, its overall performance is superior.

In Fig.~\ref{Fig_Nemenyi}, kFuse shows statistically significant differences in ARI compared to SLINK, DPC, and AP. While no significant differences in ARI are observed with other algorithms, kFuse's overall performance is better.

Fig.~\ref{Fig_Nemenyi} reveals statistically significant differences in NMI between kFuse and SLINK, DPC, and AP. Although kFuse does not show significant differences in NMI with other algorithms, its overall performance is superior.

The statistical test results confirm that the proposed kFuse algorithm outperforms all other peer algorithms (LDP-MST, SLINK, ANN-DPC, DPC, MSC, AP, K-means, and DBSCAN). This demonstrates that partitioning the dataset using natural neighbors, combined with evaluating inter-cluster similarity through boundary connectivity and density similarity, significantly improves the clustering performance of agglomerative clustering.

\section{Conclusion}

In this paper, we proposed kFuse, a density-based agglomerative clustering algorithm. kFuse performs fast, automatic, and accurate multi-prototype clustering without iteration. The entire clustering process is divided into two major stages: sub-cluster partitioning and sub-cluster merging. In the partitioning stage, kFuse uses a natural neighbor-based sub-cluster partitioning strategy, producing a reliable sub-cluster set without any user-specified parameters. In the merging stage, kFuse calculates boundary connectivity and density similarity between sub-clusters to evaluate the likelihood that two sub-clusters belong to the same cluster and merges them accordingly.

Through comparisons on synthetic and real-world datasets, as well as applications in handwritten digit and face recognition, we demonstrated the effectiveness of kFuse.

However, like K-means, kFuse requires the number of clusters to be specified beforehand. Future work aims to develop a fully automatic clustering algorithm that requires no manually specified parameters, including the number of clusters, allowing the algorithm to autonomously identify clusters within the dataset.




\bibliographystyle{IEEEtran}
\bibliography{Reference}


 





\end{document}